\documentclass[10pt,twocolumn,letterpaper]{article}

\usepackage{cvpr}              
\usepackage[accsupp]{axessibility}
\usepackage[pagebackref,breaklinks,colorlinks,allcolors=cvprblue]{hyperref}

\usepackage[capitalize]{cleveref}
\usepackage{bm}
\usepackage{comment}

\renewcommand{\vec}[1]{\boldsymbol{#1}}

\newcommand{\x}{\mathbf{x}}

\newcommand{\R}{\mathbb{R}}

\newcommand{\one}{\mathbf{1}}

\newcommand{\Rot}{\mathbf{R}}
\newcommand{\K}{\mathbf{K}}
\newcommand{\Hb}{\mathbf{H}}

\newcommand{\z}{\mathbf{z}}

\newcommand{\vb}{\mathbf{v}}

\newcommand{\bp}{\mathbf{p}}

\newcommand{\Loss}{\mathcal{L}} 

\newcommand{\pose}[0]{\vec{\theta}}
\newcommand{\wpose}[0]{\vec{^{w}\theta}}
\newcommand{\cpose}[0]{\vec{^{c}\theta}}

\newcommand{\Lsmooth}{\Loss_{\mathrm{smooth}}}

\newcommand{\Lpalm}{\Loss_{\mathrm{palm}}}
\newcommand{\Lbl}{\Loss_{\mathrm{bl}}}
\newcommand{\Lja}{\Loss_{\mathrm{ja}}}
\newcommand{\Lbio}{\Loss_{\mathrm{bio}}}
\newcommand{\Ljoints}{\mathcal{L}_{\joints}}
\newcommand{\Lshape}{\mathcal{L}_{\shape}}
\newcommand{\Lcam}{\mathcal{L}_{\mathrm{cam}}}
\newcommand{\LTwoD}{\mathcal{L}_{\mathrm{2d}}}
\newcommand{\Lpen}{\mathcal{L}_{\mathrm{pen}}}
\newcommand{\Lprior}{\mathcal{L}_{\mathrm{prior}}}
\newcommand{\Lorient}{\mathcal{L}_{\orient}}
\newcommand{\Ltrans}{\mathcal{L}_{\trans}}

\newcommand{\vid}{\mathcal{V}}
\newcommand{\Img}{\mathbf{I}}
\newcommand{\C}{\mathbf{C}}
\newcommand{\traj}{\mathbf{Q}}
\newcommand{\ctraj}{^{\mathrm{c}}\mathbf{Q}}
\newcommand{\wtraj}{^{\mathrm{w}}\mathbf{Q}}
\newcommand{\Lmat}{\mathbf{L}}
\newcommand{\handvert}{\mathbf{V}}
\newcommand{\handjoints}{\mathbf{J}}
\newcommand{\skinweight}{\mathbf{S}}
\newcommand{\hand}{\bm{q}}
\newcommand{\chand}{^{\mathrm{c}}\bm{q}}
\newcommand{\whand}{^{\mathrm{w}}\bm{q}}
\newcommand{\joints}{\bm{J}}
\newcommand{\jointsimg}{\hat{\joints}}
\newcommand{\shape}{\bm{\beta}}
\newcommand{\orient}{\bm{\phi}}
\newcommand{\trans}{\bm{\tau}}
\newcommand{\pos}{\bm{{\tau^{\mathrm{c}}}}}
\newcommand{\cjoints}{^{\mathrm{c}}\bm{J}}
\newcommand{\cshape}{^{\mathrm{c}}\bm{\beta}}
\newcommand{\corient}{^{\mathrm{c}}\bm{\phi}}
\newcommand{\ctrans}{\bm{^{\mathrm{c}}\mathrm{\tau}}}

\newcommand{\wshape}{^{\mathrm{w}}\bm{\beta}}
\newcommand{\worient}{^{\mathrm{w}}\bm{\phi}}
\newcommand{\wtrans}{\bm{^{\mathrm{w}}\mathrm{\tau}}}

\newcommand{\chandjoints}{^{\mathrm{c}}\mathbf{J}}
\newcommand{\whandjoints}{^{\mathrm{w}}\mathbf{J}}

\newcommand{\bones}{\mathbf{b}}
\newcommand{\curvs}{\mathbf{c}}
\newcommand{\dang}{\mathbf{d}}
\newcommand{\conf}{\mathbf{C}}

\newcommand{\name}{{Dyn-HaMR}}

\crefname{equation}{eq.}{eq.}
\Crefname{equation}{Eq.}{Eq.}
\crefname{theorem}{thm.}{thms.}
\Crefname{Theorem}{Thm.}{Thms.}
\crefname{conjecture}{conj.}{conjs.}
\Crefname{Conjecture}{Conj.}{Conjs.}
\crefname{proposition}{prop.}{props.}
\Crefname{proposition}{Prop.}{Props.}
\crefname{definition}{dfn.}{dfn.}
\Crefname{definition}{Dfn.}{Dfn.}
\crefname{remark}{remark}{remark}
\Crefname{Remark}{Remark}{Remark}
\Crefname{algorithm}{Alg.}{Alg.}

\crefname{section}{Sec.}{Secs.}
\Crefname{section}{Sec.}{Secs.}
\crefname{equation}{Eq.}{Eqs.}
\Crefname{equation}{Eq.}{Eqs.}
\crefname{figure}{Fig.}{Figs.}
\Crefname{figure}{Fig.}{Figs.}
\crefname{table}{Tab.}{Tabs.}
\Crefname{table}{Tab.}{Tabs.}
\crefname{thm}{Thm.}{Thms.}
\Crefname{thm}{Thm.}{Thms.}
\crefname{conj}{Conj.}{Conjs.}
\Crefname{conj}{Conj.}{Conjs.}
\crefname{dfn}{Dfn.}{Dfns.}
\crefname{dfn}{Dfn.}{Dfns.}
\crefname{remark}{remark}{remarks}
\Crefname{Remark}{Remark}{Remarks}
\crefname{prop}{Prop.}{Prop.}
\Crefname{prop}{Prop.}{Prop.}
\Crefname{algorithm}{Alg.}{Alg.}
\crefname{appendix}{App.}{apps.}
\Crefname{appendix}{App.}{Apps.}
\crefname{appsec}{appendix}{appendices}
\Crefname{appsec}{Appendix}{Appendices}

\renewcommand{\paragraph}[1]{{\vspace{1mm}\noindent \bf #1}.}

\definecolor{cvprblue}{rgb}{0.21,0.49,0.74}

\def\paperID{127} 
\def\confName{CVPR}
\def\confYear{2025}

\title{Dyn-HaMR: Recovering 4D Interacting Hand Motion from a Dynamic Camera}

\author{Zhengdi Yu  \quad\hspace{1 mm} Stefanos Zafeiriou\quad\hspace{1 mm} Tolga Birdal\\
Imperial College London\\
}

\begin{document}
\twocolumn[{
\renewcommand\twocolumn[1][]{#1}%
\maketitle
\vspace{-0.325in}
\begin{center}
    \centering
    \includegraphics[width=\textwidth]{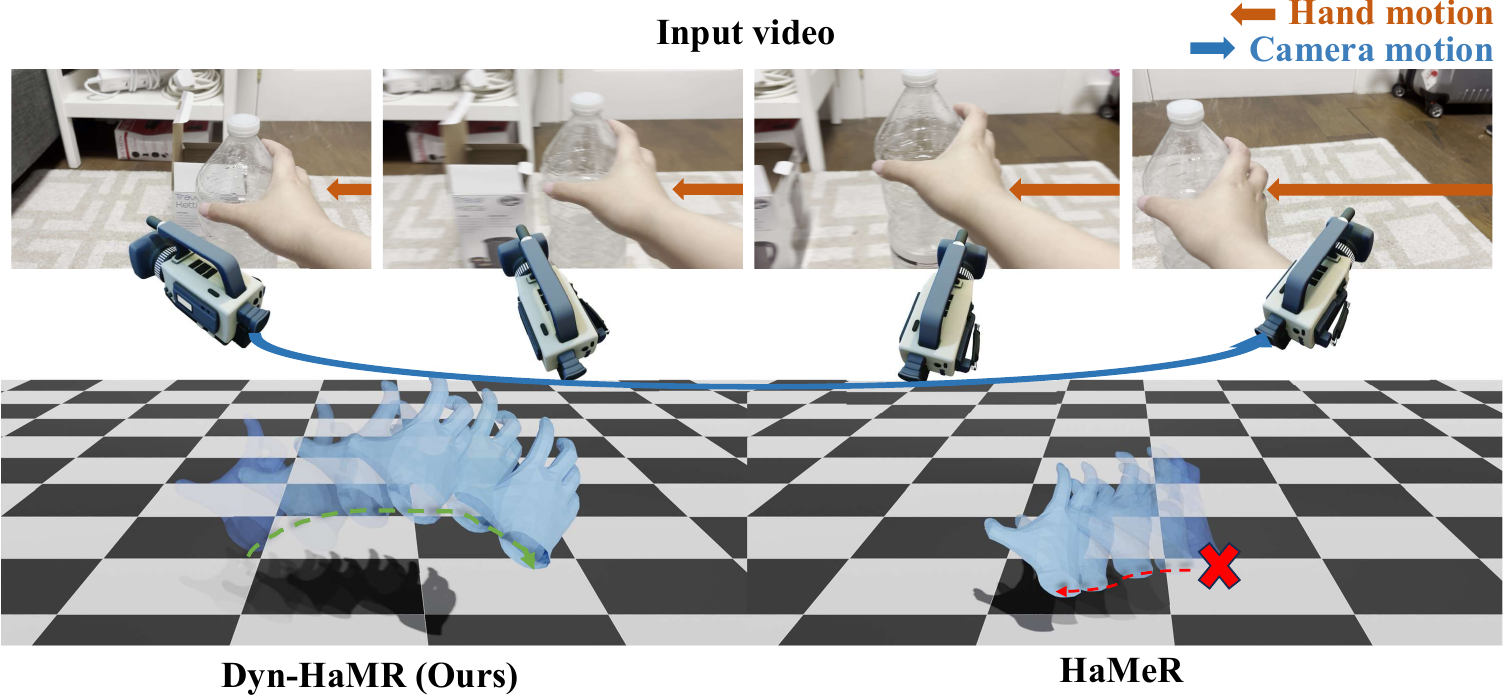}
    \captionof{figure}{\textbf{Dyn-HaMR} as a remedy for the motion entanglement in the wild. The {\textbf{green}} and {\textbf{red}} arrows represent the direction of the hand motion. \name~(Ours) can disentangle the camera and object poses to recover the 4D global hand motion in the real world whilst state-of-the-art 3D hand reconstruction methods like HaMeR \cite{pavlakos2023reconstructing}, IntagHand~\cite{Li2022intaghand} and ACR~\cite{yu2023acr} fail to do so since they cannot disentangle the sources of motion.}
    \label{fig:teaser}
\end{center}%
}]

\maketitle
\begin{abstract}
  We propose \textbf{\name}, to the best of our knowledge, the first 
  approach to reconstruct 4D global hand motion from monocular videos recorded by dynamic cameras in the wild. 
  Reconstructing accurate 3D hand meshes from monocular videos is a crucial task for understanding human behaviour, with significant applications in augmented and virtual reality (AR/VR). However, existing methods for monocular hand reconstruction typically rely on a weak perspective camera model, which simulates hand motion within a limited camera frustum. As a result, these approaches struggle to recover the full 3D global trajectory and often produce noisy or incorrect depth estimations, particularly when the video is captured by dynamic or moving cameras, which is common in egocentric scenarios. Our \name~consists of a multi-stage, multi-objective optimization pipeline, that factors in (i) simultaneous localization and mapping (SLAM) to robustly estimate relative camera motion, (ii) an interacting-hand prior for generative infilling and to refine the interaction dynamics, ensuring plausible recovery under (self-)occlusions, and (iii) hierarchical initialization through a combination of state-of-the-art hand tracking methods.
  Through extensive evaluations on both in-the-wild and indoor datasets, we show that our approach significantly outperforms state-of-the-art methods in terms of 4D global mesh recovery. This establishes a new benchmark for hand motion reconstruction from monocular video with moving cameras. Our project page is at \textbf{\url{https://dyn-hamr.github.io/}}.
\end{abstract}    
\section{Introduction}
\label{sec:intro}

In our increasingly digitalized world, capturing and interpreting human movement has become essential for advancing our interaction with computers (HCI) and immersive experiences of augmented and virtual reality (AR/VR). Many of these applications rely on a single, off-the-shelf body-mounted camera to capture hand motion, usually with complex interactions between two hands. However, as the body moves, the camera follows (\eg. egocentric), creating an intricate blend of hand and camera motions. This dynamic setup introduces a fundamental challenge: disentangling the hand motions of interest from the motion of the camera itself--a task often made difficult without auxiliary tracking. 

Most current methods for monocular hand reconstruction \cite{Li2022intaghand, yu2023acr, zhang2021interacting, moon2023interwild, lee2023im2hands, Moon_2020_ECCV_InterHand2.6M, boukhayma20193d, zhang2019end, baek2019pushing, zhou2020monocular, jiang2023a2j} assume a weak perspective camera model, capturing hand motion in either the camera coordinate frame or in a root-relative coordinate system \cite{moon2023interwild, Moon_2020_ECCV_InterHand2.6M}. These methods, however, overlook camera motion and depend solely on 2D cues, struggling with depth ambiguity, ultimately failing to recover accurate global hand trajectories.

Furthermore, hand interactions present additional obstacles, including frequent occlusions, truncation, and missed detections. Previous work either considered the simpler scenario of single-hand motion \cite{boukhayma20193d, zhang2019end, baek2019pushing, zhou2020monocular} or focused on interacting hands without specifically recovering global trajectories \cite{Li2022intaghand, yu2023acr, zhang2021interacting, moon2023interwild, lee2023im2hands}. Even with static cameras, the lack of strong interaction priors prevents existing methods from reconstructing two hands realistically under occlusions or truncations. Despite considerable progress, no approach to date has addressed the complete 4D hand reconstruction problem under the challenging conditions posed by dynamic cameras and complex hand interactions. Nor are there publicly available datasets with sufficient temporal information to enable learning of 4D global interactions.

In this work, we present \textbf{\name}: a novel, multi-stage optimization-based framework for reconstructing 4D hand motion trajectories in complex, real-world settings captured by dynamic cameras. Starting with an input RGB video, \name~leverages a robust two-hand tracking system built upon off-the-shelf methods such as MediaPipe~\cite{lugaresi2019mediapipe}, ViTPose~\cite{xu2022vitpose}, ACR~\cite{yu2023acr}, and HaMeR~\cite{pavlakos2023reconstructing} to initialize the motion state, hierarchically. We then estimate the relative camera motion using a SLAM system \cite{teed2021droid, teed2024deep}. Our multi-objective optimization ensures 3D shape projections align with 2D observations, while handling occlusions, missing detections, and ensuring plausible trajectories through a learned generative hand motion prior inspired by \cite{Duran_2024_WACV}, and augmented with biomechanical constraints. We also account for the scale factors in the hands and camera displacement, improving depth reasoning. 

Remarkably, as shown in~\cref{fig:teaser}, our approach bypasses the need for precise 3D scene reconstructions, making it adaptable to in-the-wild video data. We demonstrate our method's effectiveness through extensive experiments on dynamic, in-the-wild hand interaction videos and established benchmarks, including H2O \cite{Kwon_2021_ICCV}, EgoDexter \cite{OccludedHands_ICCV2017}, FPHA \cite{FirstPersonAction_CVPR2018}, HOI4D \cite{Liu_2022_CVPR}, and InterHand2.6M \cite{Moon_2020_ECCV_InterHand2.6M}. Our main contributions include: 
\begin{itemize} 
\item Introducing the first optimization-based approach capable of disentangling and reconstructing global 4D pose and shape of two hands, and camera trajectory. 
\item Leveraging a data-driven hand motion prior with biomechanical constraints to guide the optimization, enabling the recovery of realistic and complex hand interactions 
\item Conducting comprehensive experiments on challenging in-the-wild videos and benchmarks, demonstrating substantial performance improvements over state-of-the-art methods in 4D global motion recovery. 
\end{itemize}

Our supplementary materials provide further qualitative insights in dynamic settings using web-sourced images and videos, emphasizing the robustness of our method. We will release our implementation publicly upon publication.
\vspace{-1mm}\section{Related Work}\vspace{-2mm}
\paragraph{Monocular reconstruction of two hands} 
Pose and shape estimation of two hands (\emph{bimanual}) has rapidly progressed recently \cite{pavlakos2023reconstructing,Li2022intaghand,moon2023interwild,ren2023decoupled,yu2023acr,lee2023im2hands}.
Among them, \cite{pavlakos2023reconstructing} scaled up the single pose estimation. 
Li \etal~\cite{Li2022intaghand} developed a mesh regression network based on GCN that employs pyramid features and learned implicit attention between two hands. 
Moon el at. \cite{moon2023interwild} proposed to improve in-the-wild hand reconstruction accuracy with a strategy to bridge the domain
 gap between multi-camera datasets and in-the-wild datasets.
 Ren \etal~\cite{ren2023decoupled} used a variational autoencoder (VAE) as a prior for interacting hand reconstruction.
 Yu \etal~\cite{yu2023acr} introduced a one-stage hand reconstruction pipeline for two hands by an attention aggregation mechanism with a 2D Gaussian heatmap and cross-hand attention.
 SignAvatars~\cite{yu2024signavatars} used a multi-stage optimization pipeline to reconstruct 3D sign actors.
All these methods suffer from depth ambiguity and increased jitter in reconstruction due to the coupling of camera and hand poses. 
Moreover, reliance on weak-perspective camera models inherently limits their ability to perform effectively in dynamic scenarios.
In contrast, our optimization-based approach can reconstruct 4D global motion of two hands from complex scenes captured by a moving camera, while more accurately modelling their interaction using a learned hand motion prior.

\begin{figure*}[t]
        \centering
        \includegraphics[width=\textwidth]{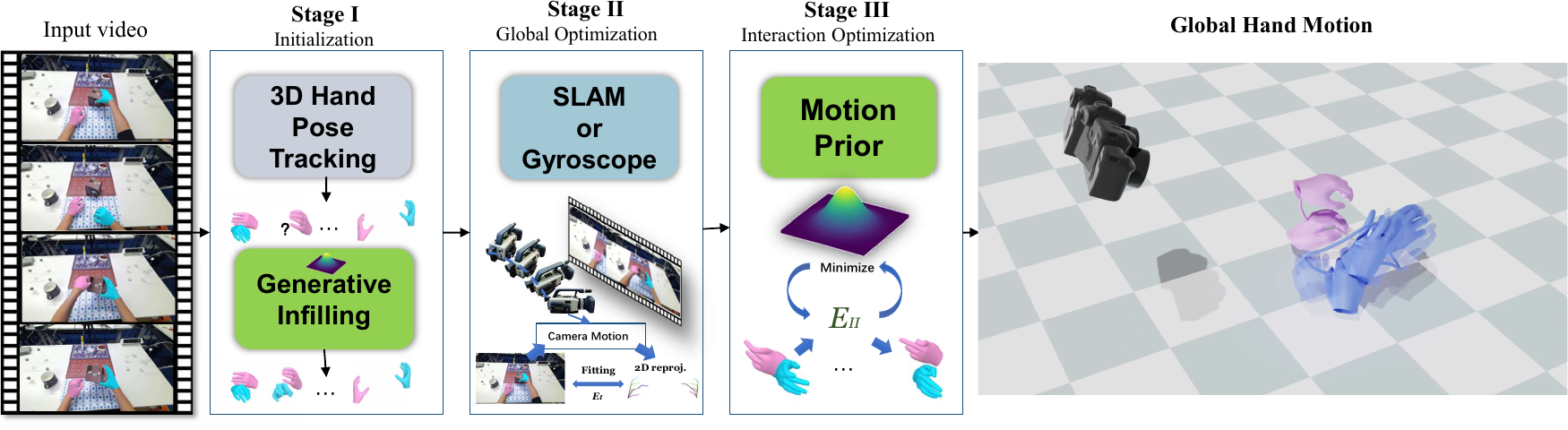}
        \caption{\textbf{Overview of our method.} We design a three-stage optimization pipeline to recover the 4D global hand motion from in-the-wild videos even with dynamic cameras. Our method can disentangle hand and camera motion as well as modelling complex hand interactions.\vspace{-3mm}}
        \label{fig:overview} 
\end{figure*}
\paragraph{4D motion priors}
Many motion models have been learned for computer animation, mostly in the context of human motion~\cite{Brand:2000:SM,Kovar:2002:MG,Rose:1998:VAA,Li:2002:MTA,Liu:2005:LPB,holden2017phase,starke2019neural}, including recent recurrent and autoregressive models~\cite{habibie2017recurrent,ghorbani2020probabilistic,henter2020moglow,yang2021real,ling2020MVAE}. These often focus on visual fidelity for a small set of characters and periodic locomotions.
Some have explored generating more general motion and body shapes~\cite{zhang2020mojo,pavllo2019modeling,aksan2019structured,Corona_2020_CVPR}, but in the context of short-term future prediction.
HuMoR~\cite{rempe2021humor} introduced an autoregressive shape prior amenable for test-time-optimization.
Other approaches tracked the full hand motion from 3D points on the surface of the hand~\cite{dewaele2004hand}. InterHandGen \cite{lee2024interhandgen} proposed a generative model as two-hand prior in close interaction with or without an object.
FourierHandFlow~\cite{lee2023fourierhandflow} devised a spatiotemporal model for 3D hand shape, using a 3D hand occupancy field with articulation-aware query flows along the temporal axis. HMP~\cite{Duran_2024_WACV} recently leveraged neural motion fields~\cite{he2022nemf} to capture the plausible space of hand movements by training on the recent Arctic~\cite{fan2023arctic} dataset. We leverage their pre-trained models to impose plausibility on the motions we recover.
\section{Dyn-HAMR}
We consider an input video $\vid=\{\Img_1,\cdots,\Img_T\}$ with $T$ frames containing two, possibly interacting hands undergoing arbitrary 6D camera motion. 
Our goal is to recover the global trajectory of both hands in the \textit{world coordinate system}. 
As shown in \cref{fig:overview}, we design a three stage optimization pipeline, inspired by the recent works in dynamic human motion perception~\cite{ye2023slahmr,shin2023wham}. The first stage (\cref{sec:3.2}) leverages the state-of-the-art interacting hand pose estimation methods~\cite{yu2023acr,pavlakos2023reconstructing,xu2022vitpose} to initialize the per-frame hand state for each of the hands in the camera coordinate system. 
Unlike human bodies, motion sequences extracted from hand images are frequently incomplete due to self-occlusions and rapid movements causing blur. As a remedy, we leverage the recent hand motion priors~\cite{Duran_2024_WACV} to perform a generative motion infilling accounting for the missed detections. In the second stage (\cref{sec:3.3}), our goal is to estimate the transformation from the world coordinate system to the camera coordinate system, while optimizing the global motion in the world coordinate system. To this end, we leverage a state-of-the-art SLAM system to compute the relative camera motions. To disambiguate the individual contributions of the camera and hand motion to the global hand motion, we also optimize for the global (world) scale factor. 
The third stage (\cref{sec:3.4}) once again leverages the learned hand motion prior, this time to further constrain the displacements of hands and refine the complex interactions together with penetration and biomechanical constraints. 


\paragraph{Representing hand motion}
We represent the global motion trajectory as a sequence of hand states, $\traj^h=\{\hand^h_t\}_{t=1}^T$.
At a given time $t$, we parameterize the hand pose and shape via MANO~\cite{romero2017embodied} model as $\hand^h_t = \{\pose^h_t, \shape^h_t, \orient^h_t, \trans_t^h \}$, where $\pose^h_t\in\R^{3\times 15}$ denotes the local hand pose in the form of $15$ hand joints, $\shape^h_t\in\R^{10}$ are the shape coefficients, and $(\orient^h_t, \trans_t^h)$ depict the global wrist pose in the form of root orientation, parameterized as the axis-angle $\orient^h_t\in\R^3$, and translation $\trans_t^h\in\R^3$. $h \in [l, r]$ determines the handedness. 
We assume the identity and the hand shape to remain unchanged through the whole sequence, \ie, $\shape^h:=\shape^h_t \,\forall\, t$.
We also use $\chand^h_t$ and $\whand^h_t$ to refer to the pose in camera and world coordinate frames, respectively. Similarly, $\ctraj^h$ and $\wtraj^h$ denote the trajectories in camera and world frames.
At a given time $t$, MANO parameters $\hand^h_t$ can be used to recover the hand mesh vertices $\handvert \in \R^{3\times 778}$ and joints $\handjoints \in \R^{3\times 21}$ through the differentiable functions: 
\begin{align}
    \handvert^h_t &= W(H(\joints^{h}_{t},\shape^{h}), P(\shape^{h}), \skinweight) + \trans^{h}_{t}\one_{778} \\
    \handjoints_{t}^{h} &=  \Lmat \handvert^h_t,
\end{align}
where $W(\cdot)$ is the skinning function, $H$ is the parametric hand template shape, and ${\one}_{778} \in \R^{1 \times 778} $ is a row vector of ones. $P$ returns the hand joint position at the rest pose, $\skinweight$ depicts the skinning weights and $\Lmat$ is a pre-trained linear regressor.

\paragraph{Hand motion priors} 
Our framework makes use of priors on hand motion both for motion infilling and for ensuring the plausibility of reconstructed sequences. Unfortunately, due to the difficulties associated with rapid hand movements and global translations, modeling a prior over hands is challenging. We make use of the only available hand prior, HMP~\cite{Duran_2024_WACV}, which is based upon neural motion fields (NeMF)~\cite{he2022nemf}. NeMF represents the motion as a continuous vector field of kinematic poses in the temporal domain by factoring out the root orientation from local motion: $D:(t,\z^h)\to\left(\joints^h_t, \left(\orient^h_t, \trans_t^h\right)\right)$, where $\z^h$ is a latent code. 
To sample novel motions, a variational framework is used where the non-autoregressive decoder $D$ maps the sampled latent vector to a spatiotemporal sequence controlled by the timestep $t$.

\subsection{Hand Tracking and Hierarchical Initialization}\label{sec:3.2}
We initialize the per-frame motion state $\hand^{h}_{t}$ by the efficient two-hand tracking system. To this end, we adopt a hierarchical initialization scheme by fusing the state-of-the-art interacting hand reconstruction methods~\cite{yu2023acr,pavlakos2023reconstructing,xu2022vitpose,lugaresi2019mediapipe}. In particular, we first fine-tune a 2D hand pose estimation model based on ViTPose~\cite{xu2022vitpose} and utilize it to obtain the bounding box sequence of each hand. Subsequently, we apply \cite{yu2023acr,pavlakos2023reconstructing,lugaresi2019mediapipe} on each cropped patch to extract a per frame motion state $\chand^{h}_t$ of each hand in the camera coordinates.

\paragraph{Motion infilling and temporal consistency}
The aforementioned single-frame interacting hand reconstruction methods naturally lack temporal coherence. Moreover, due to the frequent occlusion during hand interactions, there could be missed detection making the trajectory $^{c}Q^{h}$ \emph{incomplete}. We address both problems by employing the hand motion prior~\cite{Duran_2024_WACV} as a generative, smooth motion hallucinator. To do so, for both hands, we optimize for the latent code $\z^h$ in HMP so as to fit the frames where detections are present. We initialize this optimizer from using a canonical \emph{slerp} interpolation in the pose space. 
We compute the mean shape parameters $\shape^{h}$ based on this motion sequence, finally leading to the initial 4D hand trajectory in the camera coordinate system $\ctraj^{h}$.

We also initialize the 2D observations by incorporating ViTPose \cite{xu2022vitpose} and MediaPipe \cite{lugaresi2019mediapipe} with the reprojection of \cite{pavlakos2023reconstructing} and subsequently feed through a confidence-guided filter. 
To fill in the missed 2D keypoint detections, we reproject the 3D keypoints $
\chandjoints_{t}^{h}$ from $\chand^{h}_{t}$ onto the corresponding 2D image plane by weak-perspective camera parameters as $\mathbf{\hat{\handjoints^{h}_{t}}} \in \mathbb{R}^{3\times 21}$. This way, we get the interpolated 2D keypoints $\{\hat{\handjoints^{h}}_{t}\}^{T}_{t=0}$. For the \textit{i}-th joint at timestep $t$, $\hat{\handjoints^{h}_{t}} \in \mathbb{R}^{2\times21}$ and $\handjoints^{h}_{t} \in \mathbb{R}^{3\times21}$ have $\x=(x_{i}, y_{i}, z_{i})$ and $\bp=(\hat{x_{i}}, \hat{y_{i}})$, where the reprojection is defined as: 
\begin{equation}
    \hat{x_{i}} = sx_{i} + \tau_{x} \qquad\mathrm{and}\qquad \hat{y_{i}} = sy_{i} + \tau_{y},
\end{equation}
where $s, \tau_{x}, \tau_{y}$ are the scale and the translation for $x,y$ axes separately (\ie, the weak-perspective camera parameters estimated by the hand pose estimation model).

\subsection{4D Global Motion Optimization}\label{sec:3.3}
We now explain the recovery of the original world trajectory of interacting hands. Given the trajectory in the camera coordinate system $\ctraj$ (by \cref{sec:3.2}), our key idea here is to compute the relative camera motion with a state-of-the-art data-driven SLAM system, DPVO~\cite{teed2024deep}, and to estimate the transform $\C_{t} = \{\Rot_{t}, \pos_{t}\}$ at each timestep $t$ from the camera coordinate system to the world coordinate system. Then, the composition of hand motion $\ctraj^h$ and camera motion, \ie, $\whand^{h}_{t} = \C_{t} \odot{}\chand^{h}_{t}$ reveals the global motion. However, the scale of camera motion within the world is inherently uncertain while the hand motions are naturally constrained to be plausible. Hence, we optimize a world scale factor $\omega$ to explicitly model the relative scale between the displacements of the camera and hand motion inspired by~\cite{ye2023slahmr}.

\paragraph{Optimization variables} During optimization, we take as input the initialized 2D keypoints sequence $\{\hat{\joints}^{h}_t\}$, the 3D motion state sequence $\ctraj^h$ in the camera coordinate system, and the world-to-camera transformation $\{\Rot_{t}, \pos_{t}\}^{T}_{t=0}$ estimated by the  SLAM system, and subsequently propose a global optimization process that jointly optimizes the global trajectories, orientation and local poses of both hands and the camera extrinsics $\C_{t}$ to match the 2D observations. Specifically, we first initialize the global trajectory $\wtraj$ in the world coordinate system as follows:
\begin{align}
    \worient^h_{t} &= \Rot^{-1}_{t}{\cdot}\corient^{h}_{t}, \\
    \wtrans^{h}_t &= \Rot^{-1}_{t}  {\cdot}\ctrans^{h}_{t} - \omega \Rot^{-1}_{t}{\cdot}\pos_{t},
\end{align}
while $\wpose^{h}_{t} = {}\cpose^{h}_{t} = \pose^{h}_{t}$ and $\wshape^{h}_t = {}\cshape^{h}_t=\shape^h$ remain the same. The initial world-camera scale factor is set as $\omega=1$. With the initialized motion state $\whand_{t}^h$, the 3D mesh joints at each timestep, $\whandjoints_{t}^h$, can be extracted as below, where ${\one}_{21}\in \R^{1 \times 21} $ is a row vector of ones:
\begin{equation}
    \whandjoints_{t}^{h} = \Lmat \cdot W(H(\joints^{h}_{t} , \shape^{h}), P(\shape^{h}), \skinweight) + {}\wtrans^{h}_t\one_{21}.
\end{equation}
\paragraph{Optimization scheme} 
We recover the trajectory in the world coordinate frame by minimizing the following loss:
\begin{align}\label{eq:Eone}
    E_{I}(\whand^{h}, \omega, \Rot_{t}, \pos_{t}) = & \, \lambda_{\mathrm{2d}}\LTwoD + \lambda_{\mathrm{s}}\Lsmooth + \lambda_{\mathrm{cam}}\Lcam \notag \\
    & + \lambda_{\joints}\Ljoints + \lambda_{\shape}\Lshape.
\end{align}

The first term aligns the reprojection of $\wtraj^{h}$ with the initial 2D observations $\hat{\handjoints^{h}_{t}}$: 
\begin{equation}\label{eq:L2d}
\centering
    \LTwoD = \sum_{t=0}^{T}\sum_{h \in \{l, r\}} \rho\left(\conf^{h}_{t}\left(\Tilde{\joints_{t}^{h}}-\hat{\joints_{t}^{h}}\right)\right).
\end{equation}
Here, $\Tilde{\handjoints_{t}^{h}} = \Pi(\whandjoints_{t}^{h}, \Rot_{t}, \omega, \pos_{t}, \K)$ and $\Pi$ is the perspective camera projection with camera intrinsics $\K \in \mathbb{R}^{3 \times3}$. $\conf^{h}_{t}$ is a mask obtained from the joint visibility and $\Tilde{\handjoints_{t}^{h}}$ is the reprojected 2D keypoints from the current 3D keypoints $\whandjoints_{t}^{h}$. $\rho(\cdot)$ is the Geman-McClure robust function \cite{geman1987statistical}.

In practice, we minimize~\cref{eq:Eone} by first optimizing the root orientation $\{\orient^{h}_{t}\}^{T}_{t=0}$ and the translation $\{\trans^{h}_{t}\}^{T}_{t=0}$ in world coordinate system by minimizing \cref{eq:L2d} for 20 steps followed by 60 steps for updating the local hand pose, shape, the scale factor $\omega$,  and the camera extrinsics $\{\Rot_{t}, \pos_{t}\}^{T}_{t=0}$. 
Essentially, the motions of the two hands will further constrain the camera scale factor $\omega$ and improve the performance of complex hand interaction videos. However, the reprojection loss lacks constraints and could introduce implausible poses. To this end, we leverage the natural temporal information and introduce regularization terms for both hand and camera motion to prevent implausible poses and jittering trajectories caused by depth ambiguity:
\begin{align}
    \Lsmooth = & \sum_{t=0}^{T}\sum_{h \in \{l, r\}}\|\whandjoints_{t+1}^{h} - \whandjoints_{t}^{h}\|^{2} + d_{\pose}(\pose_{t+1}^{h}, \pose_{t}^{h})^{2} \notag \\
    \Lcam = & \sum_{t=0}^{T} d_{\Rot}(\Rot_{t+1}, \Rot_{t})^{2} + \sum_{t=0}^{T} \|\pos_{t+1} - \pos_{t}\|^{2},\nonumber
\end{align}
where rotational terms $d$ are computed as geodesic distances.
After that, in terms of reasonable regularization terms, to reduce the jittery poses, we employ standard pose $\Ljoints = \sum_{t=0}^{T} \sum_{h\in\{l, r\}}\|\joints_{t}^{h}\|^{2}$ and shape prior $\Lshape(\shape^{h}) = \sum_{h\in\{l, r\}}\|\shape^{h}\|^{2}$ term \cite{romero2017embodied}.

\subsection{Interacting Motion Prior Optimization}\label{sec:3.4}
\begin{figure}[t]
        \centering
        \includegraphics[width=\linewidth]{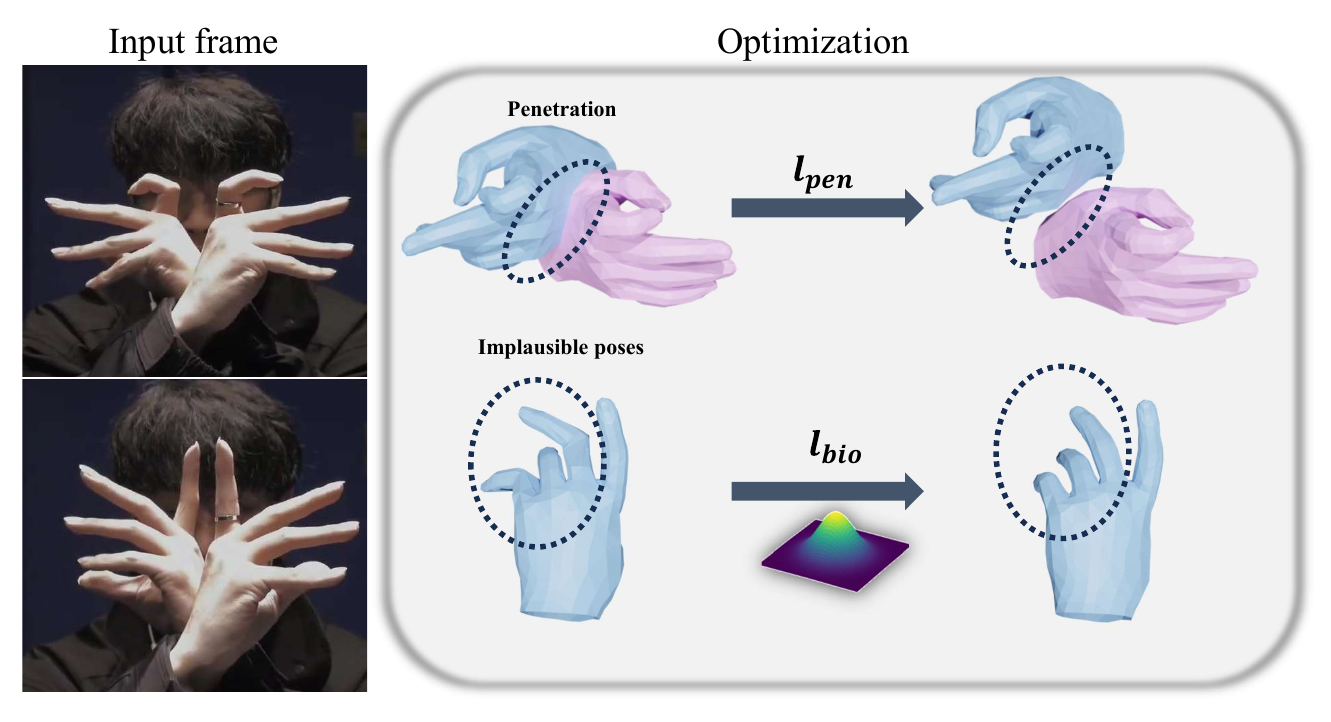}
        \caption{\textbf{Effectiveness of interaction optimization module}. Our penetration loss helps with separating hand meshes while the biomechanical constraints prevent implausible configurations that are not explained away by HMP.\vspace{-3mm}}
        \label{fig:ablation} 
\end{figure}
After the optimization of Stages I and II, we obtained global hand motion in the world coordinate system. In Stage III, we introduce an interacting hand motion prior optimization module to better model the interactions. Moreover, it constrains the displacement of the hands to be plausible, which helps to determine the contribution of the camera from the hand motion with a well-learned scale factor $\omega$. 

\paragraph{Optimization variables} For the latent optimization with motion prior \cite{Duran_2024_WACV}, we omit the decoded global orientation as it is inherently less constrained and less correlated to the local pose compared to the body pose. Specifically, we initialize the latent code $\z^{h}$ from the pre-trained encoder. Our objective is to perform the optimization over the latent code $\z^{h}$, global motion state $\wtraj^{h}$ and the scale factor $\omega$ during the optimization. Similarly, we only optimize the root orientation $\{\orient^{h}_{t}\}^{T}_{t=0}$ and the translation $\{\trans^{h}_{t}\}^{T}_{t=0}$ in the first 200 steps and add $\z^{h}$ with hand local pose and camera parameters into the optimization variables in the following 200 steps.
We recover the final interacting plausible hand motion by minimizing the following combined objective:
\begin{align}
    E_{II}(\whand^{h}, \omega, \Rot_{t}, \pos_{t}) =  &\Lprior + \Lpen + \Lbio \\
    &+ \lambda_{\mathrm{2d}}\LTwoD + \lambda_{s}\Lsmooth \nonumber\\
     &+  \lambda_{\mathrm{cam}}\Lcam + \lambda_{\joints}\Ljoints + \lambda_{\shape}\Lshape.\nonumber
\end{align}
We now define each of the sub-objectives.

\paragraph{Prior loss ($\Lprior$)}
We define $\Lprior = \lambda_{\z}\mathcal{L}_{\z} + \lambda_{\orient}\Lorient + \lambda_{\trans}\Ltrans$, where $\mathcal{L}_{\z}$ ensures that the motion is likely under the hand-motion prior by penalizing the negative log-likelihood for each of the hands:
\begin{equation}
    \mathcal{L}_{\z} = \sum_{h\in \{l ,r\}}\sum_{t=0}^{T} - \log\mathcal{N}(\\ \z^{h};\mu^{h}(\{\joints^{h}_{t}\}), \sigma^{h}(\{\joints^{h}_{t}\})).\nonumber
\end{equation}
The two other terms ensure an as jitter-free as possible trajectory by encouraging global consistency over the global orientation $\Phi^{h}$ and translation $\wtrans$:
\begin{equation}
    \Lorient = \sum_{t=0}^{T} d_{\orient}(\worient_{t},\hat{\worient_{t}}) \,\, \mathrm{and}\,\, \Ltrans = \sum_{t=0}^{T} \|\wtrans^{h}_{t} - \hat{\wtrans^{h}_{t}}\|^{2}.\nonumber
\end{equation}

\paragraph{Biomechanical loss ($\Lbio$)}
While the hand prior helps correcting certain implausible configurations, it is still necessary to explicitly constrain the hand pose for improved motion quality. Hence, we further add biomechanical constraints~\cite{spurr2020weakly} to our objective function, which consists of three terms: $\Lbio = \lambda_{ja}\Lja + \lambda_{bl}\Lbl + \lambda_{palm}\Lpalm$. For simplicity, we omit the handedness here. For $i^\mathrm{th}$ finger bone, each of the terms is defined as:
\begin{align}
    \Lja = &\sum_{i}{d_{\alpha,H}}(\bm{\alpha}_{1:T}^{i}, \Hb^{i}), \\
    \Lbl = &\sum_{i}\mathcal{I}(\|\bones^{i}_{1:T}\|_{2}; b_{\mathrm{min}}^{i}, b_{\mathrm{max}}^{i}), \\
    \Lpalm = &\sum_{i} \mathcal{I}(\|\curvs^{i}_{1:T}\|_{2}; c_{\mathrm{min}}^{i}, c_{\mathrm{max}}^{i}) \nonumber \\
          &+ \sum_{i} \mathcal{I}(\|\dang^{i}_{1:T}\|_{2}; d_{\mathrm{min}}^{i}, d_{\mathrm{max}}^{i}),
\end{align}
where $\Lbl$ is for bone length, $\Lpalm$ is for palmar region optimization, and $\Lja$ is for joint angle priors. $\Lja$ constrains the sequence of joint angles for the $i$-th finger bone $\bm{\alpha}^{i}_{1:T} = (\bm{\alpha}^{f}_{1:T}, \bm{\alpha}^{a}_{1:T})$ by approximating the convex hull on $(\bm{\alpha}^{f}_{1:T}, \bm{\alpha}^{a}_{1:T})$ plane with the point set $\Hb^i$, and the objective is to minimize the distance $d_{\bm{\alpha},\Hb}$ between them. $\mathcal{I}$ is the interval loss penalizing the outliers, and $b_i$ is the bone length of $i$-th bone. Finally, $\Lpalm$ penalizes the outliers of curvature range $(c^{i}_{\mathrm{min}}, c^{i}_{\mathrm{max}})$ and angular distance range $(d^{i}_{\mathrm{min}}, d^{i}_{\mathrm{max}})$ to constraint for the 4 root bones of palm.

\paragraph{Penetration loss  ($\Lpen$)}The final loss in this section enhances the reconstruction quality under challenging hand interactions by incorporating an interpenetration penalty to explicitly constrain the contact and the undesired penetrations between hand meshes:
\begin{equation}
\begin{split}
    \Lpen = \sum\limits_{t=0}^{T} & \left( \sum\limits_{\vb^{r}_{t} \in \handvert^{r}_{t}} 
    \min_{\vb^{l}_{t} \in \handvert^{l}_{t}} \|\vb^{l}_{t} - \vb^{r}_{t}\|^{2} \right. \\
    & \left. + \sum\limits_{\vb^{l}_{t} \in \handvert^{l}_{t}} 
    \min_{\vb^{r}_{t} \in \handvert^{r}_{t}} \|\vb^{l}_{t} - \vb^{r}_{t}\|^{2} \right).
\end{split}
\end{equation}
where $\handvert_{t}^{l}$ and $\handvert_{t}^{r}$ are the intersected vertices of the predicted left hand and right hand, respectively. 

\section{Experimental Evaluation}
We evaluate our method qualitatively and quantitatively across a variety of datasets through various metrics. 
Our suppl. material further provides additional tables, visualizations and a video demonstration.
We start by explaining our implementation, used metrics, datasets and baselines before moving onto presenting results.

\paragraph{Implementation details} We implement our method in PyTorch \cite{paszke2019pytorch}. We use the L-BFGS algorithm for our three-stage optimization with a learning rate $lr=1$. The following hyperparameters were used for each stage:
\begin{itemize}
    \item Stage II: $\lambda_{2d} = 0.001$, $\lambda_{smooth} = 10$, $\lambda_{cam}=100$ (camera displacement) and $\lambda_{\theta} = 0.04$, $\lambda_{\beta} = 0.05$.
    \item Stage III: $\lambda_{z} = 200$, $\lambda_{\phi} = 2$, $\lambda_{\gamma} = 10$, $\lambda_{pen} = 10$, $\lambda_{\beta} = 0.05$, $\lambda_{ja} = 1$, $\lambda_{palm} = 1$, $\lambda_{bl} = 1$.
\end{itemize}
For the baselines of \cite{yu2023acr,Li2022intaghand,pavlakos2023reconstructing, Duran_2024_WACV}, we adhere to their original implementation. 

\paragraph{Evaluation metrics} To thoroughly evaluate the performance of our pipeline, we conduct our evaluation on two fronts: (\textbf{i}) local pose and shape evaluation, and (\textbf{ii}) global motion reconstruction evaluation.
\begin{itemize}
    \item \textbf{Local pose and shape evaluation}: We report the Mean Per Joint Error (MPJPE), Mean Per Vertex Position Error (MPVPE) and Acceleration Error (Acc Err) measured in $mm/s^{2}$ after root alignment to evaluate the smoothness of reconstructed hand motion.
    \item \textbf{Global motion evaluation}: We quantify the errors that accumulate over time due to camera motion following \cite{yuan2022glamr, ye2023slahmr, shin2023wham}. We split sequences into 128-frame segments to optimize and align them separately with ground truth using the first two frames
    (G-MPJPE) or the whole segment (GA-MPJPE) in $mm$. 
\end{itemize}
We provide more details on implementation and evaluation in our supplementary material.

\paragraph{Datasets}
Data used for evaluation of 3D interacting hands is typically captured with static cameras, such as InterHand2.6M \cite{Moon_2020_ECCV_InterHand2.6M}, Re:InterHand \cite{moon2023reinterhand}. To assess performance under dynamic camera movement, we employ the following egocentric hand-object interaction motion datasets: 
\begin{itemize}
    \item \textbf{H2O}~\cite{Kwon_2021_ICCV} is a multiview dataset of two hands manipulating objects, which has 4 subjects performing 36 actions in 8 scenes. It provides the MANO and 3D pose annotations for both hands along with the object poses. In this work, we only focus on the egocentric view. 
    \item \textbf{HOI4D}~\cite{Liu_2022_CVPR} is a large-scale egocentric dataset that contains 4000 video sequences of 9 subjects interacting with 800 different objects from 16 categories. HOI4D provides rich annotations of panoptic segmentation, motion segmentation, 3D hand pose, object pose, and hand action. 
    \item \textbf{FPHA}~\cite{FirstPersonAction_CVPR2018} is an egocentric, dynamic, RGB-D hand dataset capturing hand in motion interacting with 3D objects, labeled with poses captured by magnetic sensors. 
    \item \textbf{EgoDexter}~\cite{OccludedHands_ICCV2017} is an RGB-D dataset for hand tracking evaluation, which contains 4 challenging occlusions and clutter scenarios with 3D joints annotation lifted from manually annotated 2D joints. 
    \item \textbf{InterHand2.6M}~\cite{Moon_2020_ECCV_InterHand2.6M} is the first publicly available interacting hands dataset with 3D mesh annotations from multiview static cameras. For our video-based experiments, we use the interacting hand (IH) subset with both human and machine annotation (H+M) of the 30 fps version.
    \item \textbf{Ego-Exo4D}~\cite{grauman2024ego} is large-scale multi-view dataset with multi-modal annotations capturing simultaneous ego and multiple exo videos. We evaluated on the subset of hand ego-pose benchmark.
    \item \textbf{HOT3D}~\cite{banerjee2024hot3d} is an egocentric dataset with multiview synchronized image streams recorded with two head-mounted devices, which provides accurate 3D poses and shapes of hands and objects.
\end{itemize}
Among them, we use the validation set for HOT3D and Ego-Ego4D datasets since their test set annotations (required to obtain GT global motion) are not publicly available this point. For other datasets used, we follow the official split to conduct our experiments.

\paragraph{Baselines} We compare our method against state-of-the-art approaches of ACR~\cite{yu2023acr}, IntagHand~\cite{Li2022intaghand} and  HaMeR \cite{pavlakos2023reconstructing} in terms of monocular hand reconstruction and global hand motion  estimation. We initialize \textbf{Ours (\name)} from \cite{pavlakos2023reconstructing} to get the results in the tables. 

\subsection{Results}
We now present our experimental results for the tasks of bimanual motion estimation from static cameras, and local and global motion estimation in dynamic settings. We also provide ablations on different stages and components.

\begin{figure*}[t]
        \centering
        \includegraphics[width=\linewidth]{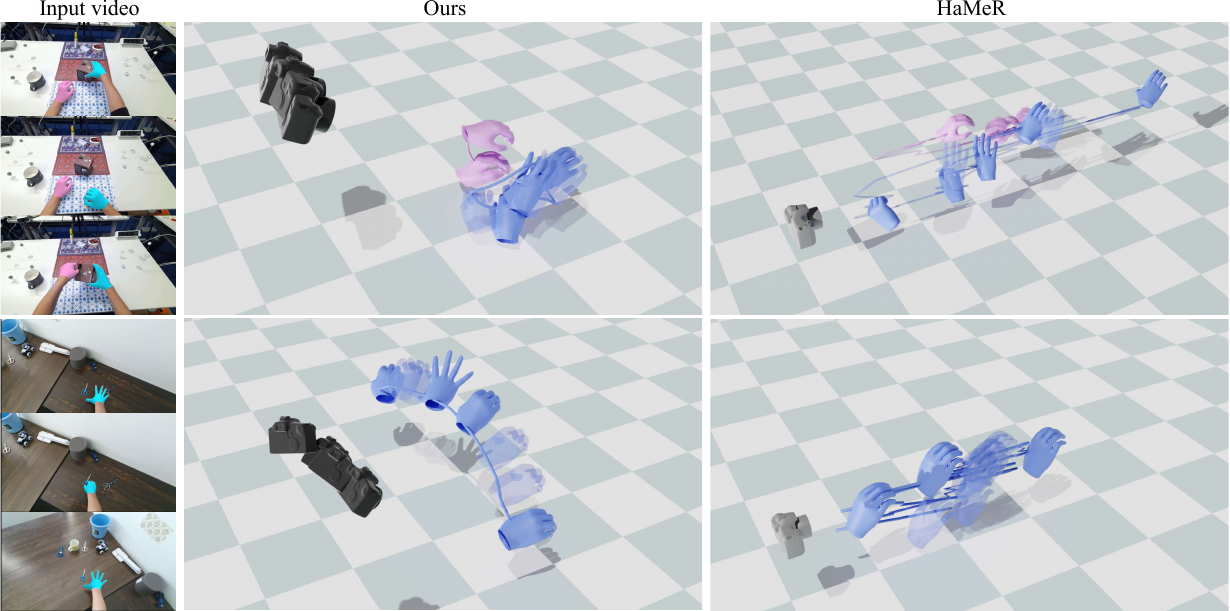}
        \caption{\textbf{Qualitative comparison with state-of-the-art method HaMeR \cite{pavlakos2023reconstructing}}. It can be seen that our method recovers significantly more plausible global hand motion. \textbf{first row} is from H2O dataset \cite{Kwon_2021_ICCV}, while the second row is from HOI4D dataset \cite{Liu_2022_CVPR}.\vspace{-3mm}}
        \label{fig:qua} 
\end{figure*}


\begin{table}[t]
\caption{\textbf{Quantitative evaluation results for InterHand2.6M \cite{moon2023reinterhand} 30 fps dataset.} We compare our method with the state-of-the-art hand reconstruction methods on local hand poses.\vspace{-6mm}}
\label{tab:interhand}
\begin{center}
\resizebox{\linewidth}{!}{
\begin{tabular}{l|ccc}
\bottomrule
\textbf{Method}               & \textbf{MPJPE} $\downarrow$ & \textbf{MPVPE} $\downarrow$ & \textbf{Acc Err} $\downarrow$  \\ 
\hline
InterWild \cite{moon2023bringing}  & 12.35 & 13.45 & 6.68 \\
DIR \cite{ren2023decoupled}  & 9.09 & 9.43 & 8.92 \\
\rowcolor{gray!7} ACR \cite{yu2023acr} & 8.75  & 9.01  & 3.99 \\
IntagHand \cite{Li2022intaghand} & 9.26 & 9.71 & 4.41 \\
HaMeR \cite{pavlakos2023reconstructing}  & 9.84 & 10.13 & 5.13 \\
\hline
Ours (w/o III)   & 8.98 & 9.25 & 4.72 \\
\rowcolor{gray!7} Ours  (\name) & \textbf{7.94} & \textbf{8.15} & \textbf{2.76} \\
\toprule
\end{tabular}
}
\end{center}
\vspace{-10mm}
\end{table}



\begin{figure}[t]
        \centering
        \includegraphics[width=\linewidth]{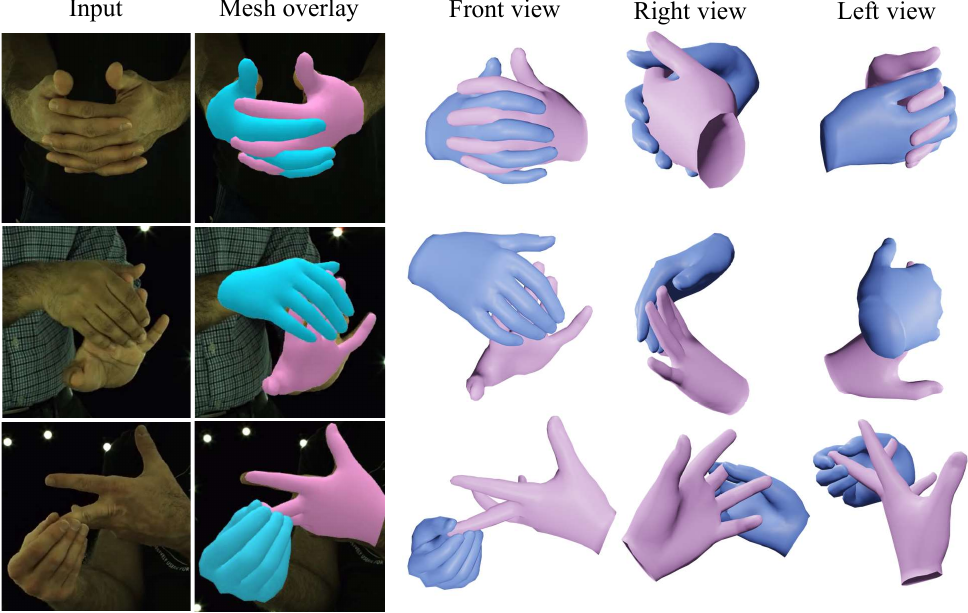}
        \caption{\textbf{Qualitative evaluation on InterHand2.6M \cite{Moon_2020_ECCV_InterHand2.6M}.} In each row, we show the mesh overlay and detailed reconstruction from different views.\vspace{-3mm}}
        \label{fig:interhand} 
\end{figure}

\paragraph{Motion estimation from static cameras}
We start by assessing our approach in the classical regime of static cameras using the 30 FPS InterHand2.6M \cite{Moon_2020_ECCV_InterHand2.6M}. \cref{tab:interhand} reports our results across all compared methods evaluated using their official configurations for a fair comparison. Despite being designed to excel in dynamic camera scenarios, our method achieves state-of-the-art performance even in the static settings, consistently surpasses existing approaches across all reported metrics.
This particularly showcases the effectiveness of our stage III, which significantly improves the plausibility of hand interactions by incorporating constraints from motion prior, biomechanical and penetration constraints, effectively mitigating depth ambiguities in the re-projection phase of earlier optimization stages. 
We present multiple views of our high-quality bimanual hand mesh reconstructions and their plausible interactions in \cref{fig:interhand}. More qualitative results including results of in-the-wild videos can be found in our suppl. material.

\paragraph{Local motion estimation}
We gauge the quality of reconstructing intricate details of the hands by measuring the joint reconstruction error, in relative to the root, a.k.a. \emph{local motion}. Results reported in \cref{tab:h2o,tab:HOI4D} show the superiority of \name~on bimanual reconstruction on H2O and HOI4D datasets, respectively, in comparison to the state-of-the-art approaches of \cite{yu2023acr,Li2022intaghand,pavlakos2023reconstructing}, where we consistently achieves the lowest MPJPE. We also assess the inter-frame smoothness and realism of motion by examining the acceleration error (Acc Err). Our approach achieves significantly lower Acc Err compared to existing methods, reflecting improved temporal consistency and smoother transitions in hand pose trajectories. Further evaluations including the FPHA dataset \cite{FirstPersonAction_CVPR2018} are included in our suppl. material.

\paragraph{Global motion estimation}
\cref{tab:h2o,tab:HOI4D} (as well as our suppl. material) reports our quantitative results on recovering local and global 4D motion under the presented metrics, using the official test split for the egocentric dynamic camera views. It can be observed that our {\name} outperforms all of the state-of-the-art methods \cite{yu2023acr,pavlakos2023reconstructing,Li2022intaghand} by a large margin in terms of \textbf{G-MPJPE} and \textbf{GA-MPJPE}. 

\cref{fig:qua} presents a qualitative comparison against state-of-the-art methods, revealing significant improvements in generating realistic and plausible motion trajectories, particularly with enhanced depth reasoning under inter-hand occlusions. Our method can better recover the actual global trajectories and respect the motion of both the hands and the moving camera, whilst other methods suffer from depth ambiguity, being agnostic to camera pose \& displacements. The entanglement of hand and camera motion further degenerates the scale estimation and lead to implausible motions. Moreover, \cref{tab:egoexo4d} and \cref{tab:hot3d} show the results on challenging two-hand object interaction scenarios, where further compare our method with the canonical HaMeR + DPVO demonstrating the vitality of our optimization. We present further qualitative results in motion reconstruction on HOI4D, FPHA and EgoDexter datasets in the suppl. material. Our suppl. video provides in-the-wild capture of hand interactions from dynamic cameras. 

\begin{table}[t]
\caption{\textbf{Quantitative evaluation results for H2O \cite{Kwon_2021_ICCV} dataset}. Our method demonstrates significant improvements over state-of-the-art approaches in recovering both local and global 4D hand motion, with additional gains achieved incorporating Stage III.\vspace{-6mm}}
\label{tab:h2o}
\begin{center}
\resizebox{\linewidth}{!}{
\begin{tabular}{l|cccc} 
\hline
\textbf{Method}               & \textbf{G-MPJPE $\downarrow$} & \textbf{GA-MPJPE $\downarrow$} & \textbf{MPJPE $\downarrow$} & \textbf{Acc Err $\downarrow$}  \\ 
\hline
ACR \cite{yu2023acr}                &             113.6           &       88.5                  &                                  46.8             &              14.3           \\
IntagHand \cite{Li2022intaghand}            &          105.5              &   81.5                      &                                   45.6            &         13.5                \\
\rowcolor{gray!7} HaMeR \cite{pavlakos2023reconstructing}                &       96.9                 &         75.7                &                                    32.9           &           9.21              \\ 
\hline
Ours (w/o III) &           51.9             &          41.2               &                                   24.9            &             9.5            \\
\rowcolor{gray!7} Ours   (\name)                &         \textbf{45.6}               &          \textbf{34.2}               &                                    \textbf{22.5}           &            \textbf{4.2}            \\
\hline
\end{tabular}
}
\end{center}
\vspace{-5mm}
\end{table}

\paragraph{Ablation studies}
To fully assess the effectiveness of \name, we perform further ablation studies on the pipeline design analyzing the contribution of each
component. In particular, we investigate the effectiveness of the key components: (i) $\mathcal{L}_{bio}$, (ii) $\mathcal{L}_{pen}$, (iii) the interacting hand motion prior module in Stage III, and (iv) generative infilling, where we replace the generative infilling module with a simple interpolation. It can be seen from \cref{tab:h2o,tab:ablation} that incorporating Stage III can boost the performance by a considerable margin as it provides a well-learned motion prior information for the final stage optimization and yields more plausible and smoother 4D trajectory reconstructions. Moreover, as shown in \cref{fig:ablation}, adding the $\mathcal{L}_{bio}$ and $\mathcal{L}_{pen}$ can significantly improve the motion quality making it more plausible and realistic.

\begin{table}[t]
\caption{\textbf{Quantitative evaluation results for HOI4D \cite{Liu_2022_CVPR} dataset.} We compare our method with the state-of-the-art hand reconstruction methods \cite{yu2023acr, Li2022intaghand, pavlakos2023reconstructing}.\vspace{-6mm}}
\label{tab:HOI4D}
\begin{center}
\resizebox{\linewidth}{!}{
\begin{tabular}{l|cccc}
\bottomrule
\textbf{Method}               & \textbf{G-MPJPE} $\downarrow$ & \textbf{GA-MPJPE} $\downarrow$ & \textbf{MPJPE} $\downarrow$ & \textbf{Acc Err} $\downarrow$  \\ 
\hline
ACR \cite{yu2023acr}                &          251.1              &   153.5                     &                                   36.4            &         12.5           \\
IntagHand \cite{Li2022intaghand}            &          291.3              &   145.6                     &                                   40.9            &         14.1                \\
\rowcolor{gray!7} HaMeR \cite{pavlakos2023reconstructing}                &          201.6              &   129.7                     &                                   27.6            &         11.6              \\ 
\hline
Ours (w/o III)  &          69.2              &       48.5                  &                                   23.7            &         10.9            \\
\rowcolor{gray!7} Ours  (\name)                &          \textbf{58.5}              &   45.6                      &                                   \textbf{19.5}            &         \textbf{4.1}            \\
\toprule
\end{tabular}
}
\end{center}
\vspace{-4mm}
\end{table}

\begin{table}[ht]
\footnotesize
\begin{center}
\caption{\textbf{Quantitative evaluation Results on Ego-Exo4D \cite{grauman2024ego} dataset.} We only report the positional error metrics as the GT annotation is too discrete to extract a meaningful acc.}\label{tab:egoexo4d}
\vspace{-3mm}
\setlength{\tabcolsep}{3pt}
\resizebox{\linewidth}{!}{
\begin{tabular}{l|ccc} 
\bottomrule
\textbf{Method} & \textbf{Jerk $\downarrow$} & \textbf{PA-MPJPE $\downarrow$} & \textbf{G-MPJPE $\downarrow$} \\ 
\hline
\rowcolor{gray!7} ACR & 245.12 & 19.45 & 291.34 \\ 
IntagHand & 175.65 & 24.32 & 275.89 \\ 
HaMeR & 195.45 & 15.87 & 267.75 \\ 
HaMeR + DPVO & 200.45 & 15.87 & 213.75 \\ 
\hline
\rowcolor{gray!7} Ours (\name)  & \textbf{5.26} & \textbf{14.34} & \textbf{53.89} \\ 
\toprule
\end{tabular}
}
\end{center}
\vspace{-2mm}
\end{table}

\begin{table}[ht]
\footnotesize
\begin{center}
\caption{\textbf{Qualitative results on HOT3D \cite{banerjee2024hot3d} dataset}. We compare our method with the canonical HaMeR + DPVO baseline.}\label{tab:hot3d}
\vspace{-3mm}
\setlength{\tabcolsep}{3pt}
\resizebox{\linewidth}{!}{
\begin{tabular}{l|cccc} 
\bottomrule
\textbf{Method} & \textbf{Jerk $\downarrow$} & \textbf{PA-MPJPE $\downarrow$} & \textbf{G-MPJPE $\downarrow$} & \textbf{Acc Err $\downarrow$} \\ 
\hline
\rowcolor{gray!7} ACR & 153.45 & 16.41 & 159.34 & 16.45 \\ 
IntagHand & 171.24 & 21.75 & 165.42 & 15.12 \\ 
HaMeR & 189.62 & 10.43 & 155.98 & 13.78 \\ 
HaMeR + DPVO & 195.77 & 10.43 & 129.45 & 12.78 \\ 
\hline
\rowcolor{gray!7} Ours (\name)  & \textbf{4.18} & \textbf{8.87} & \textbf{42.36} & \textbf{4.95} \\ 
\toprule
\end{tabular}
}
\end{center}
\vspace{-2mm}
\end{table}

\begin{table}[t]\vspace{-2mm}
\label{tab:ablation}
\caption{\textbf{Ablation of pipeline components on H2O \cite{Kwon_2021_ICCV} dataset.} It shows the impact of removing different components from the pipeline on various performance metrics.\vspace{-6mm}}
\begin{center}
\resizebox{\linewidth}{!}{
\begin{tabular}{l|cccc} 
\bottomrule
\textbf{Method}              & \textbf{G-MPJPE $\downarrow$} & \textbf{GA-MPJPE $\downarrow$} & \textbf{MPJPE $\downarrow$} & \textbf{Acc Err $\downarrow$}  \\ 
\hline
Stage I                 &         84.5               &          72.5               &                                    25.6           &            8.8            \\
Stage I+II                 &           51.9             &          41.2               &                                   24.9            &             9.5            \\
\hline   
w/o bio. const. &   49.6      &    43.1      &      24.5          &  4.3        \\
w/o pen. const. &    46.3     &       34.7          &   23.6    &  \textbf{4.1}        \\
w/o gen. infill.               & 48.9  &  37.8  & 24.1  &  5.6 \\
\rowcolor{gray!7} Ours (\name)                &         \textbf{45.6}               &          \textbf{34.2}               &                                    \textbf{22.5}           &            4.2            \\
\hline
\end{tabular}
}
\end{center}
\vspace{-4mm}
\end{table}



\section{Conclusion}
We introduced \textbf{\name}, to the best of our knowledge, the only data-driven work, which could reliably recover the 4D global motion of two interacting hands from complex, in-the-wild videos containing complex scenes, acquired by moving, dynamic cameras. \name~achieves this by leveraging a state-of-the-art SLAM system in conjunction with proposed interacting hand priors. 
Our method consists of a multi-objective optimization pipeline in which we estimate the relative camera motions and address motion entanglement as well as depth ambiguity problems by aligning the reconstruction with 2D observations. We further incorporate a learnable world scale factor to disambiguate the contributions of local hand and camera motions to the global hand motion.
Our interaction priors allow for filling in the missing detections while ensuring plausible hand trajectories as demonstrated with extensive evaluations. 

\paragraph{Limitation \& future work} 
While \name~successfully works for limited time-horizons, extending it to long sequences with generative, extrapolation capabilities remains to be explored. Developing a regression-based method is one of the possible directions to deal with long sequences. We will also work on improving the hand priors extending~\cite{he2024nrdf} and incorporate object interactions.



\clearpage
\section*{Acknowledgments}
T. Birdal and S. Zafeiriou acknowledge support from the Engineering and Physical Sciences Research Council [grant EP/X011364/1].
T. Birdal was supported by a UKRI Future Leaders Fellowship [grant number MR/Y018818/1] as well as a Royal Society Research Grant RG/R1/241402.

{
    \small
    \bibliographystyle{ieeenat_fullname}
    \bibliography{main}
}

\clearpage
\newpage
\appendix
\section*{Appendices}

\setcounter{page}{1}

The following sections supplement our main paper in terms of implementation details and additional qualitative and quantitative evaluations. Moreover, we refer the reader to our \textcolor{Rhodamine}{\textbf{supplementary videos}} on the project webpage to better perceive the resulting {motions} and for a more comprehensive exposition, where we show promising results on in-the-wild hand reconstruction and qualitative comparisons against the state-of-the-art interacting hand reconstruction methods \cite{pavlakos2023reconstructing} under the challenging dynamic camera scenarios. 

\section{Implementation Details and Data Pre-Processing}
We now provide more details into our initialization before moving onto the optimization scheme.  

\subsection{Initializing Motion States in Camera Frame}
To initialize 2D observations in image plane and the MANO parameters $\chand^h_t = \{\pose^h_t, \shape^h_t, \corient^h_t, \ctrans_t^h \}$ in the camera coordinate system at timestep $t$, we adopt a hierarchical pipeline. 

First, we employ a 2D hand pose estimation model, ViTPose, following \cite{xu2022vitpose, pavlakos2023reconstructing}, known for its performance in hand detection and palm localization. Despite its strength in detecting global hand regions, the model often produces jittery and inaccurate joint positions, making it insufficient for subsequent optimization processes. As a remedy, we refine the 2D inputs by cropping the image based on bounding boxes calculated from ViTPose's 2D keypoint predictions. Specifically, for a set of keypoints $^{\mathrm{vit}}\jointsimg^{h}_{t}$, the bounding box is calculated by the point sets with a confidence filter $\epsilon_{b} = 0.5$ and an extension coefficient of $200\%$. To initialize the MANO parameters $\{\pose^{h}_{t}, \shape^{h}_{t}, \corient^{h}_{t}\}$, we run the state-of-the-art hand reconstruction method of \cite{pavlakos2023reconstructing}, using the officially released checkpoints, on the image patches restricted to the calculated bounding box. 
To estimate translation $\ctrans^{h}_{t}=(x,y,d)$ of the two hands in 3D space, where $d$ is the direction along depth, we simulate various versions of $\ctrans^{h}_{t}$ in the camera coordinate system based on the predicted weak-perspective camera parameters $(s, t_{x}, t{y})$, assuming a fixed focal length $f=1000$ following \cite{pavlakos2023reconstructing} using $x = t_{x}, y = t_{y}, d = \frac{2f}{s \times s_{\Img}}$, where $s_{\Img}$ is the image size. Alternatively, the camera translation in the camera coordinate system can also be acquired by solving the PnP algorithm (\ie RANSAC \cite{fischler1981random}) with the 3D keypoints $\cjoints^{h}_{t}$ and their corresponding 2D projections $\jointsimg^{h}_{t}$ on the image plane. Finally, we infill the missed frames if the interval is less than 50 frames using the approach described in Sec. \textcolor{cvprblue}{3.1}.

\paragraph{Keypoint refinement} Building upon the complete 3D initialization, we refine the corresponding 2D observations to improve accuracy and consistency. Specifically, we first detect all hands in the scene using ViTPose \cite{xu2022vitpose}, and then combine these detections with predictions from MediaPipe \cite{lugaresi2019mediapipe} and the 2D re-projections derived from the 3D initialization. To achieve this, we extract wrist positions from ViTPose and pair them with finger joint predictions from MediaPipe, ensuring that both hands in each pair are correctly matched to the same individual. We replace the 2D finger joints whose confidence scores are lower than the threshold $\epsilon_{j} = 0.5$ with the corresponding 2D re-projections from the 3D initialization.

\paragraph{Handling occlusions} Modelling accurate interactions between hands is particularly challenging due to frequent occlusions, rapid motions, and truncations. These factors often lead to missed detections, especially in complex interaction scenarios. To address this, we employ a generative motion infilling approach, as detailed in Sec. \textcolor{cvprblue}{3.1}. Specifically, we infill the hands from the timestep where it first appears to the last appearance timestep $(t_{start}, t_{end})$ with our generative motion prior. To handle the missed detections and occlusion more robustly, we only optimize the visible individual hands and mask out the objective terms for the occluded frames (\ie we only update the latent code $\z$ with these observed timesteps in Stage III), utilizing the motion prior as a guide to reason and infill the occluded frames.

\paragraph{Handling hallucinations} HaMeR can yield erroneous hallucinations -- such as multiple hands in the same location, incorrect handedness, or implausible poses.
Specifically, HaMeR's detector can produce overlapping bounding boxes (bboxes) without suppression, leading to redundant or inconsistent predictions as both ViTPose and HaMeR process each bbox independently.
As a remedy, we use ViTPose to extract 2D keypoints with confidences (IoU $>0.9$) and instead of processing all overlapping detections, we retain only the bbox with the highest confidence before feeding it into HaMeR. 
We also filter any detection that appears in $<$10 frames, reducing false positives.
Incorrect handedness, where one hand is occasionally confused as its opposite, can be identified by the sudden change in the bbox (IoU with the previous bbox $<$0.1), allowing us to mark these frames as invalid prior to generative infilling.

\subsection{Optimization Scheme} 
\paragraph{Multi-stage optimization} Our key insight is to optimize the interacting hands in stages, balancing the per-frame motion accuracy and temporal smoothness while avoiding over-smoothing We first optimize the two hands during Stage II individually with a lower $\lambda_{smooth}=1$ to ensure accurate local pose and pixel alignment.  After obtaining plausible global motion, we start to jointly optimize two hands in a single batch with interacting hand motion prior module, which makes the scale information shared between the two hands and further constrains the hands-camera displacement plausible. During optimization, the dimension of the latent code $\z$ is 128 in the hand motion prior module. Interpenetration is only applied when both hands are present in the scene.

\paragraph{Chunk optimization} (i) For the pre-processing of long sequence $\vid=\{\Img_1,\cdots,\Img_T\}$ with $T \geq 128$ frames, we segment each video into chunks of 128 frames. This ensures compatibility with the hand motion prior module, which adopts a sequence length of 128 for motion parameterization as per \cite{Duran_2024_WACV}. Subsequently, we optimize the motion segments in chunks. We initialize the next motion and camera state with the end state of the last chunk (\eg initialize $\C_{127}$ and $\hand^h_{128}$ with the optimized output of $\hand^h_{127}$ and $\C_{128}$), as well as the world scale factor $\omega$. (ii) In terms of the post-processing for evaluation, we align the translation parameters across segments and combine them to generate seamless visualizations of the reconstructed motion.

\section{Evaluation Metrics} To compute \textbf{G-MPJPE and GA-MPJPE}, we first align the first two frames (G-MPJPE) or the whole sequence (GA-MPJPE) of hand motion with the GT using Umeyama method. We then transform the prediction to align with the GT before computing the MPJPE as the mean $L_2$ distance between each predicted and GT joint. 
To compute the \textbf{Acc Err}, we followed the common practice in HMP, GLAMR and SLAHMR ignoring the division, $\alpha_{i}^{t} = v^{t-1}_i - 2v^t_i + v^{t+1}_i$, where $\alpha_{i}^{t}$, $v^{t-1}_i$ are the acc. and velocity at timestep $t$ without the division of  discretized time step. 
We further analyze the effect of the different computation of acceleration where the discretized time step is applied or not. We report both Acc Err with and w/o division ($mm^{2}$ and $m/s^{2}$) in \cref{tab:acc}, where we can observe that our method keeps consistently better results under different computation method.
\begin{table}[t]
\vspace{-5mm}
\centering
\caption{\textbf{Acceleration analysis on HOT3D dataset.} Acc Err is reported w/o the div. of $\omega^{2}$ (left) in $mm/s^{2}$ and with the div. of $\omega^{2}$ in $m/s^{2}$ (right). Lower ($\downarrow$) is always better.}\label{tab:acc} 
\resizebox{\linewidth}{!}{\begin{tabular}{l|cc}
\hline
Method & Acc Err (w/o) $\downarrow$ & Acc Err (with) $\downarrow$ \\
\hline
ACR \cite{yu2023acr} & 16.45 & 14.82 \\
IntagHand \cite{Li2022intaghand} & 15.12 & 13.62 \\
HaMeR \cite{zuo2023reconstructing} & 13.78 & 12.41 \\
\hline   
HaMeR + DPVO \cite{teed2024deep} & 12.78 & 11.51 \\
\rowcolor{gray!7} Ours (Dyn-HaMR) & \textbf{4.95} & \textbf{4.46} \\
\hline
\end{tabular}}
\end{table}

\noindent
For the \textbf{RTE} ($\%$) of sequence with $N$ frames, we compute it the as the Trajectory Error as in \cref{sec:Additional_Experiments}: ${\frac1{N} \sum_{i=1}^{N} \| \mathbf{T}_{\text{target}}^{i} - (\mathbf{R} \cdot \mathbf{T}_{\text{pred}}^{i} + \mathbf{t}) \|_2}/{\Delta}$, where $\Delta=\sum_{i=1}^{N-1} \| \mathbf{T}_{\text{target}}^{i+1} - \mathbf{T}_{\text{target}}^{i} \|_2$, with $(\mathbf{R},\mathbf{t})$ being the computed rigid transform and $\mathbf{T}$ the root translation.

\begin{figure*}[t]
        \centering
        \includegraphics[width=1\linewidth]{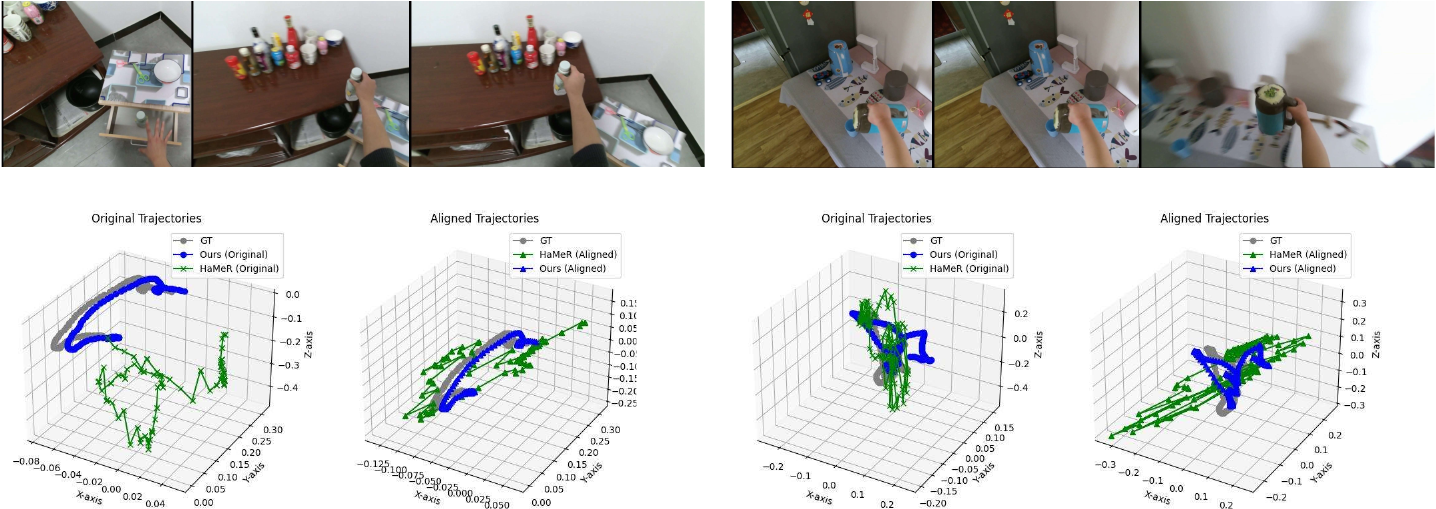}
        \caption{\textbf{Comparison of global trajectory on HOI4D} \cite{Liu_2022_CVPR}.}
        \label{fig:supp_trans} 
\end{figure*}
\section{Additional Experiments}\label{sec:Additional_Experiments}
In this section, we provide experiments for our 4D global hand motion reconstructions from both in-the-wild videos and existing benchmarks (\ie H2O \cite{Kwon_2021_ICCV}, FPHA \cite{FirstPersonAction_CVPR2018}, HOI4D \cite{Liu_2022_CVPR}, EgoDexter \cite{OccludedHands_ICCV2017}). 

\paragraph{Evaluation metrics} We evaluate both the reconstruction quality and the plausibility of our motion. In addition to the evaluation metrics of \textbf{(i)} local hand pose and shape, \textbf{(ii)} global hand motion, we further conduct \textbf{(iii) motion plausibility evaluation} quantifying the plausibility and fidelity of our bimanual reconstructions. In addition to the metrics introduced in our main paper such as MPJPE ($mm$), PA-MPJPE ($mm$) and Acc Err ($mm/s^{2}$), we further propose the following two fronts for evaluation:
\begin{itemize}
    \item \textbf{Global trajectory plausibility:} We quantify Trajectory Error (Trans Err) in \% for each clip after the rigid alignment and normalize it by displacements of ground truth trajectory.
    \item \textbf{Bimanual (interacting hand) pose plausibility:} We report Fréchet Distance (FID) between estimations and the GT data to quantify the plausibility of the joint pose of interacting hands. To evaluate the smoothness and the interaction quality, we compute Jerk ($10m/s^{3}$) in the world coordinate system and Mean Inter-penetration Volume (Pen) in $cm^3$ to measure between the two hands.
\end{itemize}
Specifically, we adopt a PointNet++ \cite{qi2017pointnet++} based embedding network for Fréchet distance on latent space following \cite{shu20193d, lee2024interhandgen}. For single-hand plausibility, we train it on the combined dataset of InterHand2.6M and H2O to regress the local hand poses $\pose \in\R^{3\times 15}$ in axis-angle representation from the hand mesh vertices $\handvert \in \R^{3\times 778}$. We keep the original setting while only modifying the last layer. The reconstruction MPVPE achieves $1.18mm$. For interacting hands, we modify the last layer to regress the hand pose for both hands as well as the relative root translation and rotation. This achieves $1.65mm$ and $1.61mm$ MPVEP for the left and right hands, respectively. We report the FID score for both the single hand version and the bimanual version, whenever two hands are jointly visible.  

\begin{figure*}[t]
        \centering
        \includegraphics[width=1\linewidth]{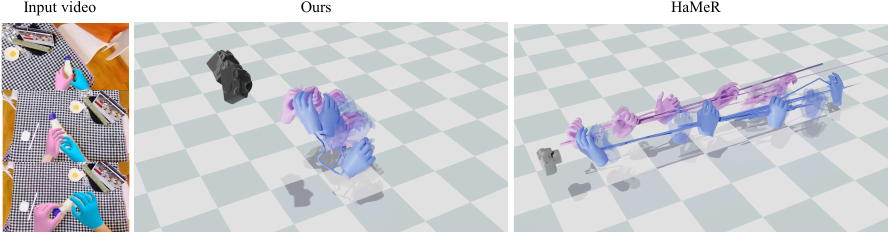}
        \caption{\textbf{Qualitative comparison with state-of-the-art method HaMeR \cite{zuo2023reconstructing} on in-the-wild online videos.}}
        \label{fig:momo} 
\end{figure*}

\subsection{Results}
\begin{figure*}[t]
        \centering
        \includegraphics[width=1\textwidth]{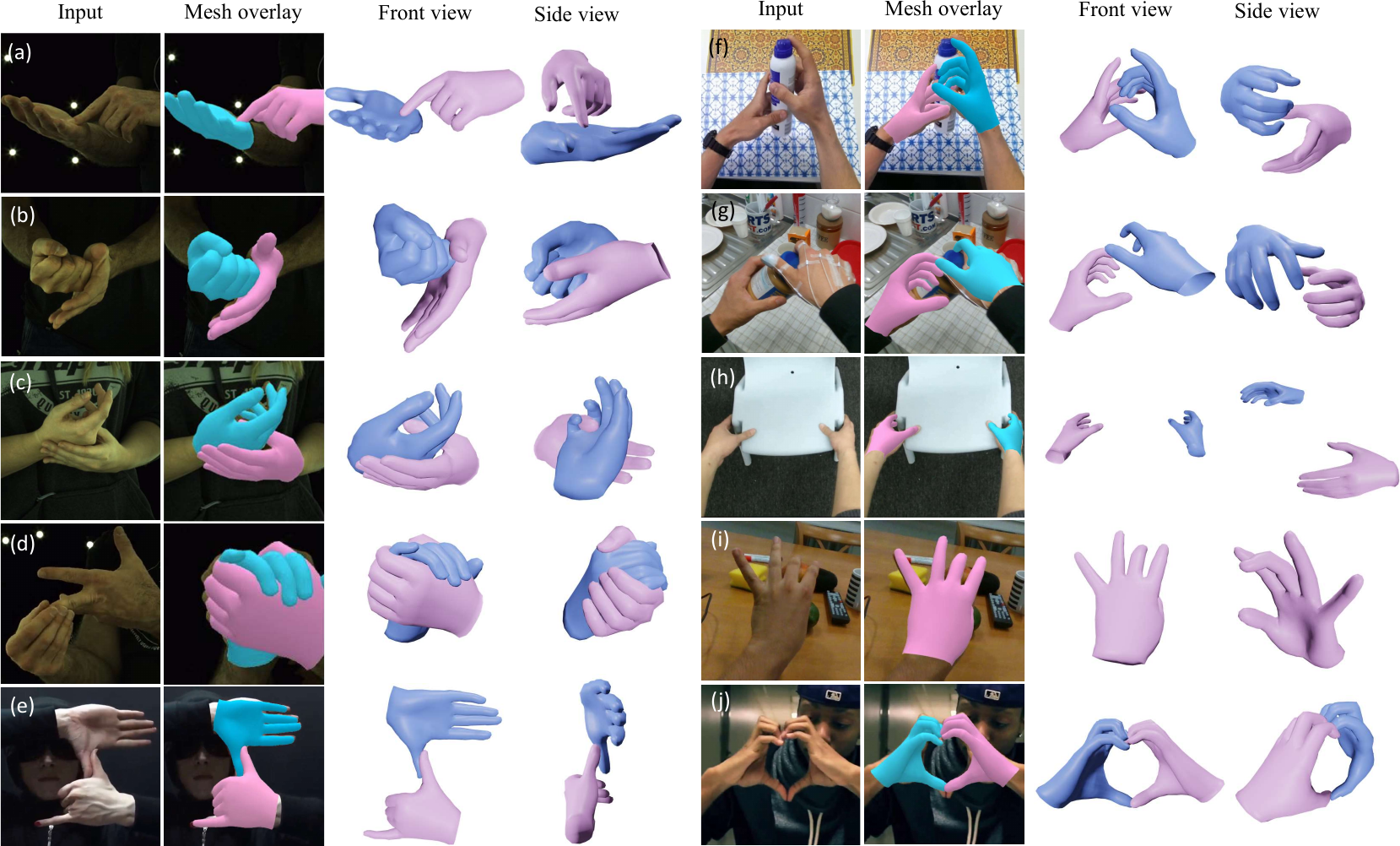}
        \caption{\textbf{Qualitative results of hand motion estimation under complex hand interactions and hand-object interaction.} In the figure, (a)-(b) show the samples from InterHand2.6M dataset \cite{Moon_2020_ECCV_InterHand2.6M}, while (f)-(i) are from H2O \cite{Kwon_2021_ICCV}, FPHA \cite{FirstPersonAction_CVPR2018}, HOI4D \cite{Liu_2022_CVPR} and EgoDexter \cite{OccludedHands_ICCV2017}, respectively. Finally, (e) and (j) are reconstruction results from in-the-wild web videos.\vspace{-3mm}}
        \label{fig:local} 
\end{figure*}

\paragraph{Local motion estimation} To fully analyze the effectiveness of our pipeline, we conduct experiments on \textbf{FPHA} \cite{FirstPersonAction_CVPR2018}, an egocentric RGB-D hand-object motion dataset, which contains 105K frames encompassing 45 daily hand action categories, captured across diverse hand configurations with ground truth 3D hand joint annotations provided in the camera coordinate system. We evaluate the quality of bimanual hand reconstruction by measuring the root-relative joint reconstruction error (local motion) using metrics such as MPJPE, PA-MPJPE, and Acc Err. As shown in \cref{tab:fpha}, our method demonstrates superior performance, achieving the lowest MPJPE (18.9 mm) and PA-MPJPE (12.5 mm) compared to other state-of-the-art approaches. Additionally, we achieve a competitive Acc Err of \textbf{5.7}, demonstrating improved temporal smoothness and consistency. Furthermore, qualitative results in \cref{fig:local} further illustrate the robustness of our pipeline in handling complex in-the-wild scenarios such as egocentric hand-object interactions and hand-hand interactions.

\begin{figure}[t]
        \centering
        \includegraphics[width=0.5\textwidth]{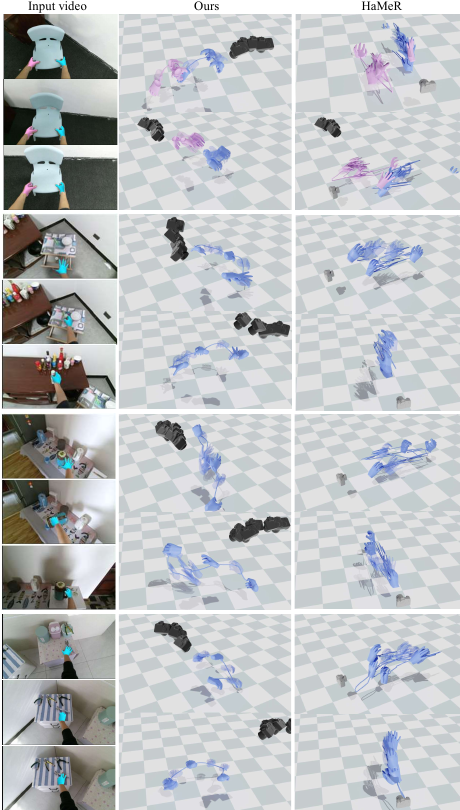}
        \caption{\textbf{In-the-wild 4D global hand motion reconstructions on HOI4D dataset \cite{Liu_2022_CVPR}.} We visualize the front view and the bird’s eye view, which are the upper row and the lower row in each of the sample motions. State-of-the-art hand reconstruction approach HaMeR \cite{pavlakos2023reconstructing} fails to recover plausible global trajectories while our method produces. Moreover, our method produces significantly less jitter and more plausible depth reasoning. Please see the \textcolor{Rhodamine}{\textbf{supplementary video}} for better visualization of motions.}
        \label{fig:global} 
\end{figure}

\begin{figure*}[t]
        \centering
        \includegraphics[width=0.9\textwidth]{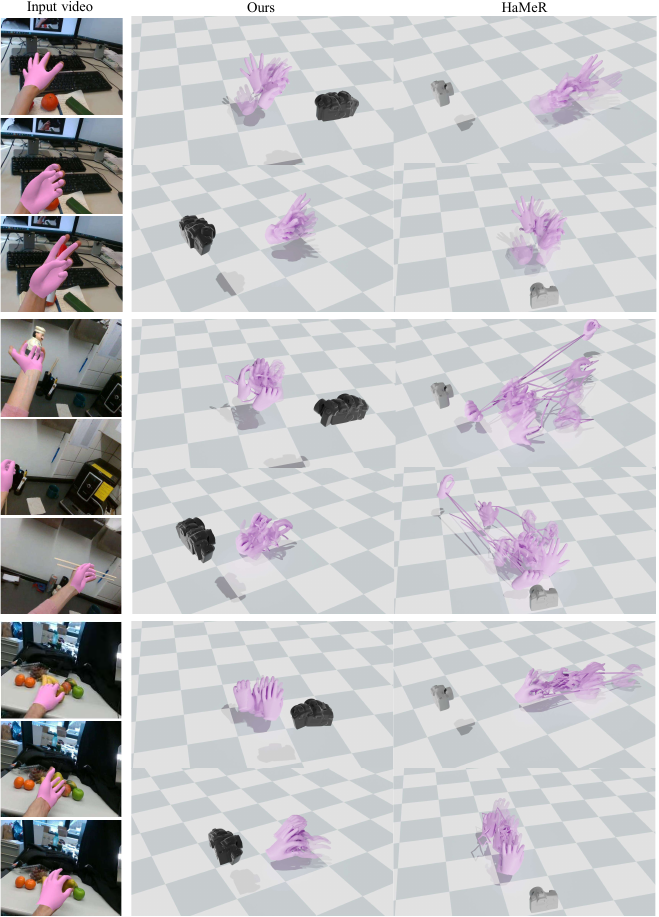}
        \caption{\textbf{In-the-wild 4D global hand motion reconstructions on EgoDexter dataset \cite{OccludedHands_ICCV2017}}. Please see the supplementary video for the motion visualization. Our method produces pausible global motion and depth reasoning.}
        \label{fig:egodexter} 
\end{figure*}

\begin{figure*}[t]
        \centering
        \includegraphics[width=0.9\textwidth]{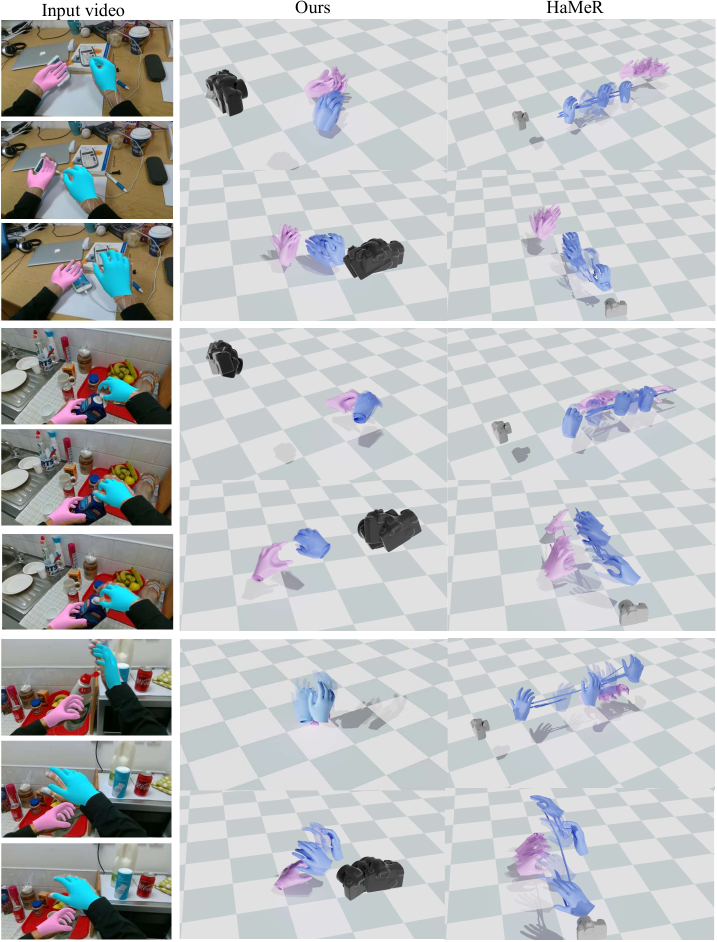}
        \caption{\textbf{In-the-wild 4D global hand motion reconstructions on FPHA dataset \cite{FirstPersonAction_CVPR2018}}. We also provide detailed motion visualization of FPHA in the supplementary video.}
        \label{fig:fpha} 
\end{figure*}

\paragraph{Global motion estimation} As introduced in Sec. \textcolor{cvprblue}{4.1} of the main paper, H2O \cite{Kwon_2021_ICCV} and HOI4D \cite{Liu_2022_CVPR} contain dynamic camera videos with available camera poses to convert the hand poses from the camera coordinate system to the world coordinate system. To evaluate global motion recovery performance, we have conducted qualitative evaluations on the aforementioned four egocentric hand-object interaction datasets mentioned in Sec. \textcolor{cvprblue}{4.1} as well as on in-the-wild videos. In this section, we first present qualitative results on these datasets shown in \cref{fig:global,fig:fpha,fig:egodexter}, where our method produces plausible 4D global motion with trajectories in the world coordinate system, while previous state-of-the-art methods fail to capture the global motion in 3D space, especially from dynamic cameras. Furthermore, our method yields more plausible depth reasoning in the bimanual setting and significantly reduces the jitter in translations. To quantify the reconstruction accuracy and errors, we evaluate the Translation Error and Jerk in \cref{tab:plausibility}, where we can observe significant improvements over the state-of-the-art approaches. Specifically, we evaluate the RTE score for each of the motion trajectories in the world coordinate system. It can be observed that our method consistently archives the lowest translation error across datasets. We also conduct further analysis on the HOI4D dataset as shown in \cref{fig:supp_trans}, which contains large camera displacements. Notably, we achieve 3.89\% in Trans Err against 18.98\% of HaMeR \cite{pavlakos2023reconstructing} on this dataset. We encourage to browse the reconstruction results in our \textbf{supplementary videos}, which clearly demonstrates the superiority of our pipeline against the state-of-the-art methods.

\begin{figure*}[t]
        \centering
        \includegraphics[width=1\textwidth]{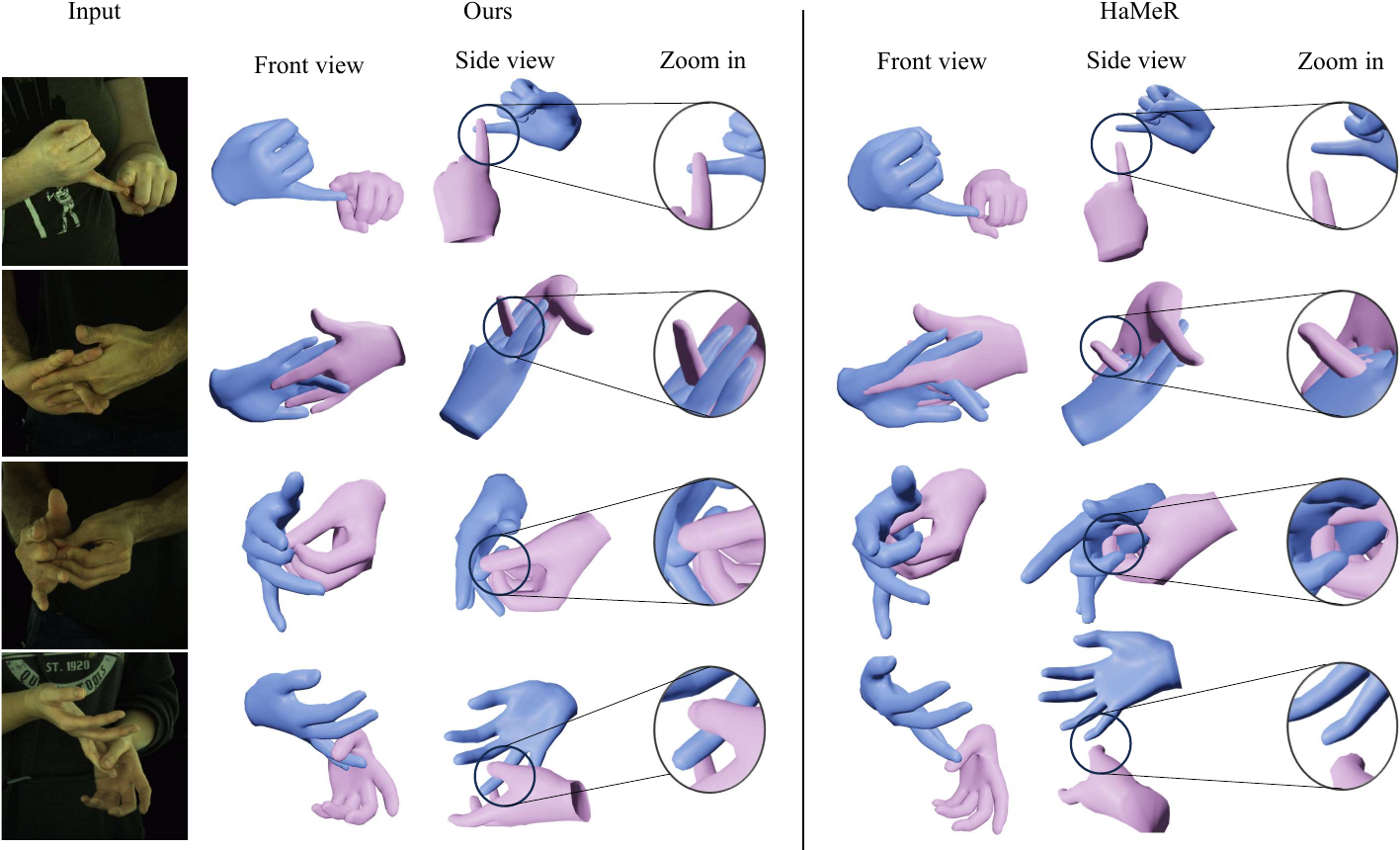}
        \caption{\textbf{Comparison with state-of-the-art hand reconstruction approach (static camera) on InterHand2.6M dataset \cite{Moon_2020_ECCV_InterHand2.6M}.} We compare our method with state-of-the-art hand reconstruction approach HaMeR \cite{zuo2023reconstructing} under challenging hand interactions.}
        \label{fig:compare_static} 
\end{figure*}

\begin{figure}[t]
        \centering
        \includegraphics[width=0.5\textwidth]{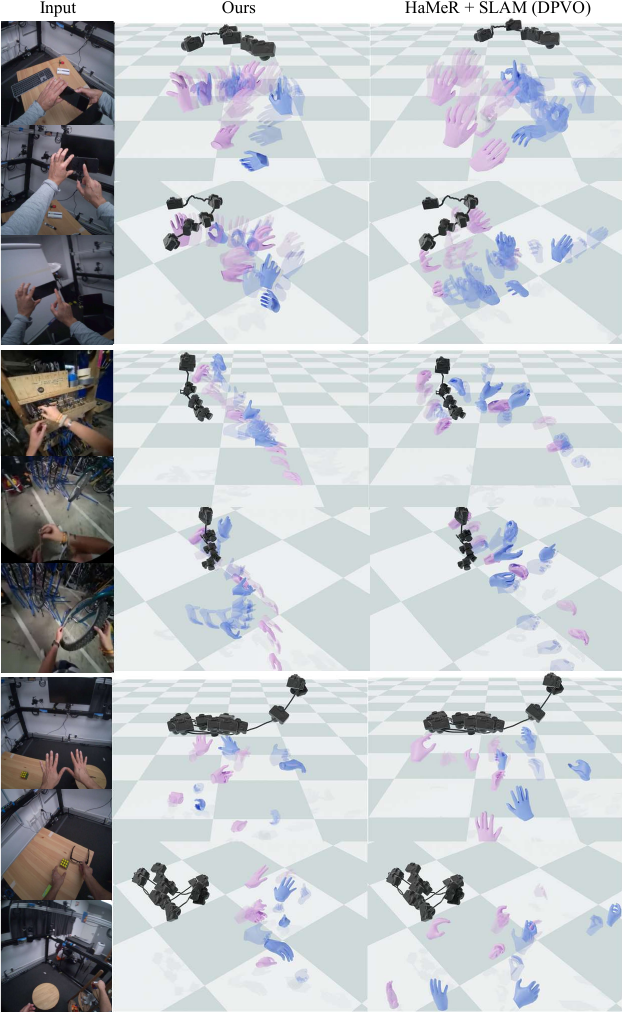}
        \caption{{In-the-wild 4D global hand motion reconstructions on {HOT3D}~\cite{banerjee2024hot3d}  (top and bottom rows) and {Ego-Exo4D}~\cite{grauman2024ego} (middle row)  datasets.}}
        \label{fig:wild} 
\end{figure}

\begin{table}[t]
\caption{\textbf{Quantitative comparison on FPHA \cite{FirstPersonAction_CVPR2018} dataset.} PA-MPJPE represents the MPJPE after Procrustes Alignment. \vspace{-6mm}}
\begin{center}
\resizebox{\linewidth}{!}{
\begin{tabular}{l|ccc} 
\bottomrule
\textbf{Method}               & \textbf{MPJPE $\downarrow$} & \textbf{PA-MPJPE $\downarrow$} & \textbf{Acc Err $\downarrow$} \\ 
\hline
ACR \cite{yu2023acr}   & 43.6 & 35.1 & 13.1 \\
IntagHand \cite{Li2022intaghand}    & 41.2 & 31.6 & 12.4 \\
\rowcolor{gray!7} HaMeR \cite{pavlakos2023reconstructing}   & 29.9 & 18.7 & 12.5 \\ 
\hline   
w/o bio. const. & 19.6 & 13.5 & 6.1 \\
w/o pen. const. & 21.3 & 15.7 & \textbf{5.4} \\
\rowcolor{gray!7} Ours (\name)     & \textbf{18.9} & \textbf{12.5} & 5.7 \\
\hline
\end{tabular}
}\label{tab:fpha}
\end{center}
\vspace{-3mm}
\end{table}

\paragraph{Bimanual hand pose plausibility}
\begin{table*}[ht]
\footnotesize
\begin{center}
\caption{\textbf{Plausibility evaluation on multiple datasets.} Results are reported on the H2O \cite{Kwon_2021_ICCV} and InterHand2.6M \cite{Moon_2020_ECCV_InterHand2.6M} to analyze the jitter, penetration, translation, and plausibility. FID is reported for both single hand (left) and two hands (right).\vspace{-3mm}}\label{tab:plausibility}
\resizebox{\linewidth}{!}{
\begin{tabular}{l|cccc|cccc} 
\bottomrule
\textbf{Method} & \multicolumn{4}{c|}{\textbf{H2O}} & \multicolumn{4}{c}{\textbf{InterHand2.6M}} \\ 
\cline{2-9}
 & \textbf{Jerk $\downarrow$} & \textbf{Pen $\downarrow$} & \textbf{Trans Err $\downarrow$} & \textbf{FID $\downarrow$} & \textbf{Jerk $\downarrow$} & \textbf{Pen $\downarrow$} & \textbf{Trans Err $\downarrow$} & \textbf{FID $\downarrow$} \\ 
\hline
\rowcolor{gray!7} ACR \cite{yu2023acr} & 149.43 & 0.07 & 10.89 & 1.95 / 4.45 & 153.62 & 5.05 & 8.65 & 2.51 / 5.36 \\ 
IntagHand \cite{Li2022intaghand} & 166.38 & 0.06 & 11.15 & 2.14 / 4.12 & 165.31 & 4.82 & 9.19 & 2.69 / 5.07 \\ 
HaMeR \cite{zuo2023reconstructing} & 195.77 & 0.06 & 10.43 & 1.76 / 4.78 & 183.45 & 5.17 & 8.43 & 2.45 / 5.45 \\ 
\hline
Ours (w/o bio. const.) & 2.65 & 0.04 & 4.71 & 1.89 / 2.78 & 4.57 & 2.67 & 4.41  & 1.89 / 4.12 \\ 
Ours (w/o pen. const.) & 2.36 & 0.02 & 4.13 & 1.38 / 2.12 & \textbf{4.03} & 4.23 & 4.93 & 1.53 / 4.64 \\ 
Ours (w/o III) & 2.98 & 0.02 & 4.21 & 2.01 / 2.93 & 4.81 & 4.49 & 4.96 & 2.89 / 4.87 \\ 
\rowcolor{gray!7} Ours ({\name}) & \textbf{2.34} & 0.009 & \textbf{5.67} & \textbf{1.34} / \textbf{1.98} & 4.26  & \textbf{2.46} & \textbf{4.35} & \textbf{1.49} / \textbf{3.56} \\ 
\toprule
\end{tabular}
}
\end{center}\vspace{-6mm}
\end{table*}
In addition to evaluating global motion recovery, we conduct extensive experiments on complex interacting hand scenarios and assess the plausibility of the results. \cref{fig:compare_static} indicates significantly more details in reconstruction in favor of our method, more stable \textbf{depth reasoning}, and higher local hand pose accuracy under self-occlusions compared to the baseline \cite{pavlakos2023reconstructing}. We further provide a plausibility evaluation of the 4D motion reconstructions in \cref{tab:plausibility}, where our approach is compared against state-of-the-art methods \cite{yu2023acr, pavlakos2023reconstructing, Li2022intaghand} and ablations of different modules. Our method consistently outperforms existing approaches by a large margin across all reported metrics. Thanks to the integration of penetration and biomechanical constraints, our approach exhibits superior stability in recovering bimanual poses from complex interacting scenarios, achieving the lowest FID and Jerk. This demonstrates the effectiveness of our Stage III in addressing the challenges of bimanual interactions.


\subsection{Analysis}
\paragraph{Ablation study of initialization} While our network is not restricted to any specific initialization backbone we provide an additional ablation study on the 3D motion state initialization to fully assess the effect of each component, where we conduct experiments on ACR \cite{yu2023acr}, IntagHand \cite{Li2022intaghand} and HaMeR \cite{pavlakos2023reconstructing}. As illustrated in \cref{tab:ablation_supp}, we compare our full pipeline (HaMeR \cite{pavlakos2023reconstructing} initialization) with initializations from \cite{yu2023acr,Li2022intaghand} and Ours (Base) represents to initialize from the default MANO mean pose. It can be observed that there is a boost in the performance with the recent large-scale model based hand reconstruction framework HaMeR \cite{pavlakos2023reconstructing}, which indicates that a better initialization could further improve the optimization and speed up the convergence.

\begin{table}[hbt]
\footnotesize
\begin{center}
\caption{\textbf{Ablation study on H2O \cite{Kwon_2021_ICCV} dataset.} To quantify the importance of the initialization, we compare the performance initialized from different state-of-the-art approaches \cite{yu2023acr,Li2022intaghand,pavlakos2023reconstructing}.\vspace{-2mm}}\label{tab:ablation_supp}
\resizebox{\linewidth}{!}{
\begin{tabular}{l|cccc} 
\bottomrule
\textbf{Method}               & \textbf{G-MPJPE $\downarrow$}  & \textbf{GA-MPJPE $\downarrow$} & \textbf{MPJPE $\downarrow$} & \textbf{Acc Err $\downarrow$} \\
\hline
Ours (Base) &  55.8    &   47.6      &  28.9  &   4.2      \\
\hline
Ours (ACR \cite{yu2023acr}) &  49.8    &   37.3   &   23.2  &   4.7      \\
Ours (IntagHand \cite{Li2022intaghand}) &   48.9      &    41.4      &      25.1          & 4.5       \\
\rowcolor{gray!7} Ours (HaMeR \cite{zuo2023reconstructing}) &         \textbf{45.6}               &          \textbf{34.2}               &                                    \textbf{22.5}           &            \textbf{4.2}            \\
\hline
Ours (Long)                 &   69.5      &    49.1      &          22.3    &   4.2       \\
\toprule
\end{tabular}
}
\end{center}\vspace{-7mm}
\end{table}

\paragraph{Runtime} Our network is agnostic to the initialization method (e.g. camera, hand initialization), which affects the processing time. As shown in \cref{tab:runtime}. On an NVIDIA A100 GPU, our experiments for a 128-frame video clip adopt HaMeR and ACR for 3D initialization, taking 3.18 minutes, and 8 seconds on average, respectively. Subsequently, optimizing stage II takes around 2.3 minutes. Finally, the last stage takes 1 to 2 minutes. We use Pyrender for off-screen rendering, which takes an additional minute. Note, the rendering time can also vary depending on the specific resolution and number of views desired. Compared to existing optimization-based pipelines such as humor \cite{rempe2021humor} and slahmr \cite{ye2023slahmr}, which take more than 45 minutes to 2 hours for 128 frames on A100 GPU, our method achieves the fastest test time optimization, which makes a step towards efficient and real-time applications.

\paragraph{Long sequences degeneration} As described in Sec. \textcolor{cvprblue}{1} and \textcolor{cvprblue}{3}, the errors in estimated global trajectories would accumulate over time in our moving camera setting. Therefore, we follow standard evaluations for open-loop reconstruction (\eg, SLAM and inertial odometry) to compute errors using a sliding window, similar to WAHM and GLAMR. To quantify its impact, we provide the results for $long$ sequence here in \cref{tab:ablation_supp}, where we conduct the evaluation based on the original video sequence length instead of the 128 clips mentioned in the experiments section.


\begin{table}[t]
\centering
\caption{\textbf{Runtime}. We show the individual runtime for initialization and each optimization stage separately.\vspace{-2mm}}
\label{tab:runtime}
\resizebox{\linewidth}{!}{\begin{tabular}{l|c}
\hline
\textbf{Methods}                & \textbf{Avg. runtime (min.)} \\ \hline
HuMoR \cite{rempe2021humor}           & 58.7 \\ 
SLAHMR \cite{ye2023slahmr}            & 65.5 \\ 
\hline
Camera tracking (DPVO \cite{teed2024deep})         & 1.49 \\ 
3D Hand tracking (HaMeR \cite{zuo2023reconstructing})           & 3.18 \\ 
3D Hand tracking (ACR \cite{yu2023acr})           & 0.13 \\ 
2D keypoints detection \cite{lugaresi2019mediapipe,xu2022vitpose}   & 1.45 \\ 
Stage II optimization        & 2.5 \\ 
Stage III optimization          & 1.69 \\ \hline
\textbf{{\name} (initialization)}                  & 3.07$\sim$6.12 \\ 
\textbf{{\name} (optimization)}                  & 4.19 \\ 
\hline
\end{tabular}}
\vspace{-3mm}
\end{table}

\end{document}